\documentclass[10pt]{OpenEnvision}

\usepackage{amsmath,amssymb,textcomp,array,longtable,tabularx}
\usepackage{tikz}
\usetikzlibrary{arrows.meta,positioning,calc}
\usepackage{algorithm,algorithmic}
\usepackage{url,xspace,multicol}
\colorlet{tableaccent}{oered!65!oeink}
\colorlet{tableheadfill}{oered!9!white}
\definecolor{oursfill}{HTML}{F1F3F5}
\newcommand{\tabletoprule}{\arrayrulecolor{tableaccent}\toprule[0.6pt]\arrayrulecolor{black}}
\newcommand{\tablehead}{\rowcolor{tableheadfill}}

\newcommand{\oursrow}{\rowcolor{oursfill}}
\newcommand{\best}[1]{{\bfseries #1}}
\newcommand{\second}[1]{\underline{#1}}
\newcolumntype{L}[1]{>{\raggedright\arraybackslash}p{#1}}
\newcolumntype{Y}{>{\raggedright\arraybackslash}X}
\newcolumntype{R}{>{\raggedleft\arraybackslash}X}
\DeclareRobustCommand{\sysname}{\mbox{WM$^\text{MLLM}$}\xspace}

\graphicspath{{figures/}}
\makeatletter\g@addto@macro{\UrlBreaks}{\UrlOrds}\makeatother
\pretocmd{\subsection}{\par\ifinner\vspace{12pt}\fi}{}{\PackageError{main}{Subsection spacing patch failed}{Inspect the heading command.}}

\ifXeTeX\microtypesetup{protrusion=false}\fi
\setlist[itemize]{nosep,topsep=2pt,partopsep=0pt}
\setlist[enumerate]{nosep,topsep=2pt,partopsep=0pt}
\titleformat{\section}{\sectionfont\sffamily\bfseries\color{oered}}{\thesection}{0.7em}{}[\nobreak\vspace{2pt}\color{seedpanelborder}\nobreak\titlerule]
\titlespacing*{\section}{0pt}{11pt plus 2pt minus 2pt}{5pt}
\titlespacing*{\subsection}{0pt}{8pt plus 2pt minus 1pt}{3pt}
\titlespacing*{\subsubsection}{0pt}{6pt plus 1pt minus 1pt}{3pt}
\titlespacing*{\paragraph}{0pt}{3.5pt plus 1pt}{0.6em}
\DeclareCaptionStyle{ruled}{format=custom,labelsep=custom,font=small}

\fancypagestyle{reportstyle}{%
  \fancyhf{}
  \fancyhead[L]{\sffamily\bfseries\small\color{oeink}\sysname Technical Report}
  \fancyhead[R]{\sffamily\bfseries\small\color{oered}\sysname}
  \fancyfoot[C]{\sffamily\small\color{oemuted}\thepage}
  \renewcommand{\headrulewidth}{0.5pt}
  
}
\makeatletter\let\ps@plain\ps@reportstyle\makeatother
\title{\textcolor{oered}{\sysname}: Turning MLLMs into World Models}
\author[1,3]{Baiang Li}
\author[2,\dagger]{Enze Xie}
\author[2]{Jincheng Yu}
\author[2]{Jin Wang}
\author[2]{Haozhe Liu}
\author[2]{Yitong Li}
\author[3]{Juanxi Tian}
\author[1]{Felix Heide}
\contribution[\dagger]{{\protect\affiliationfamily Project Lead}}
\renewcommand{\affiliationlist}{%
  \begin{minipage}[c]{0.30\linewidth}\centering
    \includegraphics[height=11.5mm,trim=0 0 153.25bp 0,clip]{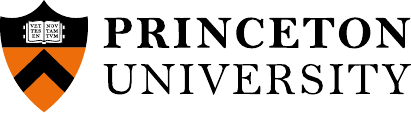}\par
    \vspace{1.5mm}{\affiliationfont\affiliationfamily $^1$Princeton University}\par
  \end{minipage}\hfill
  \begin{minipage}[c]{0.30\linewidth}\centering
    \includegraphics[height=10.5mm,trim=0 0 1096.84bp 0,clip]{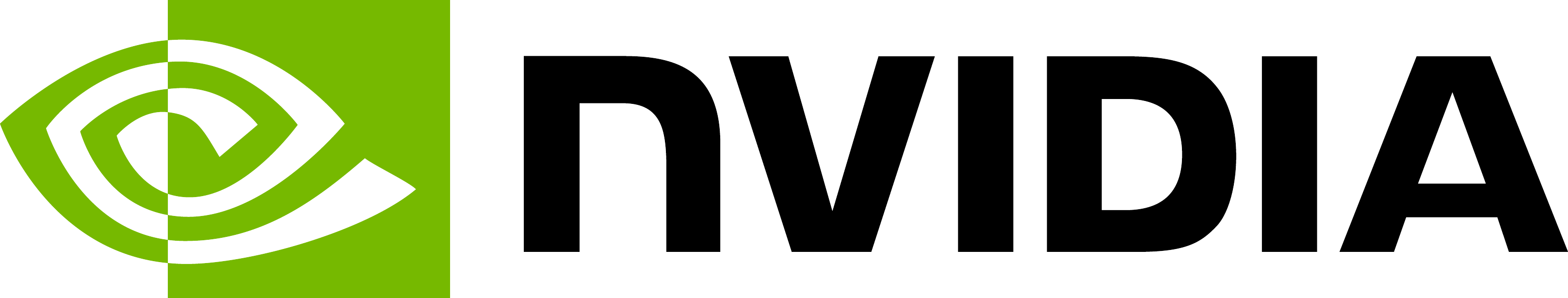}\par
    \vspace{1.5mm}{\affiliationfont\affiliationfamily $^2$NVIDIA}\par
  \end{minipage}\hfill
  \begin{minipage}[c]{0.30\linewidth}\centering
    \includegraphics[height=11.5mm]{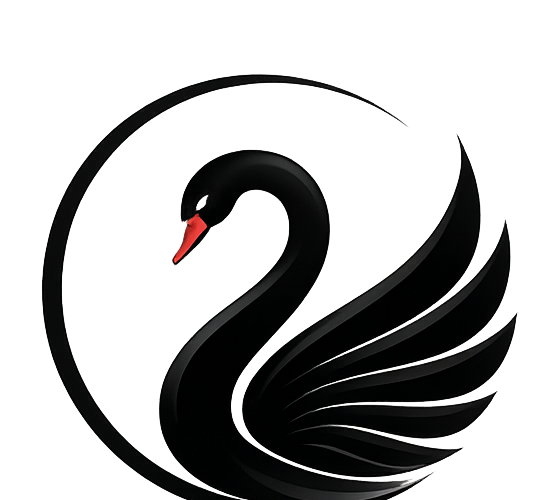}\par
    \vspace{1.5mm}{\affiliationfont\affiliationfamily $^3$OpenEnvision}\par
  \end{minipage}%
}
\checkdata[Project page]{\url{https://ztmotalee.github.io/wmmllm/}}
\abstract{Video world models keep the world in the context of a generator, so a revisit is synthesized again and earlier edits can drift or disappear. We present \sysname{}, the first interactive world model built on multimodal large language models (MLLMs), for persistent, editable world generation from a single image. \sysname{} keeps the world state outside every generator, as a panorama with a revision history, and lets the MLLMs write it: they turn each request into operations on this state, select where a local edit is written, and check every candidate before it becomes the next state. Image generators only propose views, and video only presents them. Users can walk to new places and return, edit objects through text or a box, place subjects, and change lighting, weather or season. By construction, a revisit at the same revision reproduces the earlier view exactly, a local edit leaves every panorama pixel outside its region unchanged, and each request starts from the result of the previous one. Against six world model baselines on MBench-A, \sysname{} has the highest mean on all four environment-recovery metrics (Epipolar 91.5 versus 0.1 to 74.0) and completes the most camera round trips (85.3\% versus at most 50.9\%). It has the best value in all 15 CompBench columns against 15 baselines. On 112 local edits judged blind, selecting the region to write after generation raises completion from 34.8\% to 83.0\%. On 258 WBench and MBench-A items, all 46 revisits recover their earlier views, and all 97 local edits that pass the check leave panorama pixels outside their region unchanged.
}

\hypersetup{hypertexnames=false,pdftitle={WM\textasciicircum MLLM: Turning MLLMs into World Models},pdfauthor={Baiang Li, Enze Xie, Jincheng Yu, Jin Wang, Haozhe Liu, Yitong Li, Juanxi Tian, Felix Heide}}

\begin{document}
\clubpenalty=10000
\makeatletter\@clubpenalty=\@M\makeatother
\widowpenalty=10000
\displaywidowpenalty=10000
\newgeometry{top=16mm,bottom=25mm,left=25mm,right=25mm}
\maketitle
\setcounter{tocdepth}{1}
\begingroup
\small
\setlength{\columnsep}{8mm}
{\cfttoctitlefont\contentsname\par}\vspace{1pt}
\makeatletter
\begin{multicols}{2}
\@starttoc{toc}
\end{multicols}
\makeatother
\endgroup
\restoregeometry
\pagestyle{reportstyle}

\section{Introduction}
\label{sec:intro}
\begin{figure}[t!]
\centering
\includegraphics[width=\textwidth]{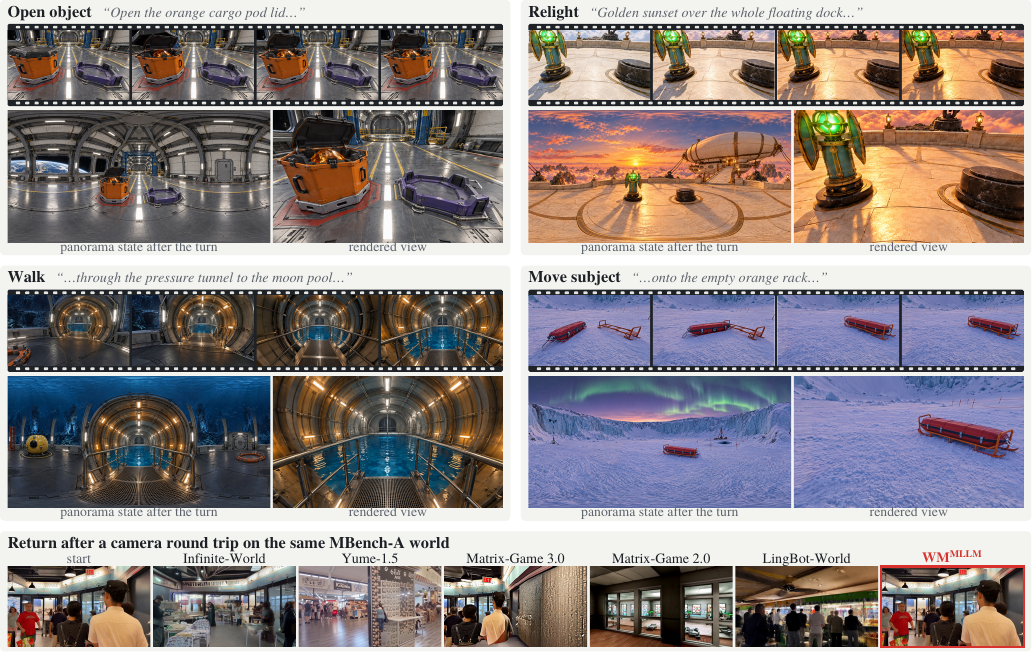}
\caption{\textbf{Overview of \sysname{} results.} \sysname{} opens objects, relights scenes, walks to new places and moves subjects; each panel shows one turn: video frames, the panorama state and the rendered view. After one camera round trip, five baselines end on different scenes and \sysname{} on the starting view (mean pixel difference between the first and last frame, 0--255: 88.4 to 101.3 against 1.1).}
\label{fig:teaser}
\end{figure}

A user opens a container, relights the room and turns away. Turning back should show the open container under the new light (\cref{fig:teaser}). A world model keeps a persistent state of its environment that actions update~\cite{worldfoundry}, so an interactive world model needs three properties: a revisit shows the content saved there (\emph{return}), a local edit changes nothing outside its region (\emph{locality}), and every request builds on the result of the previous one (\emph{persistence}).

\begin{figure}[tbp]
\centering
\includegraphics[width=\textwidth]{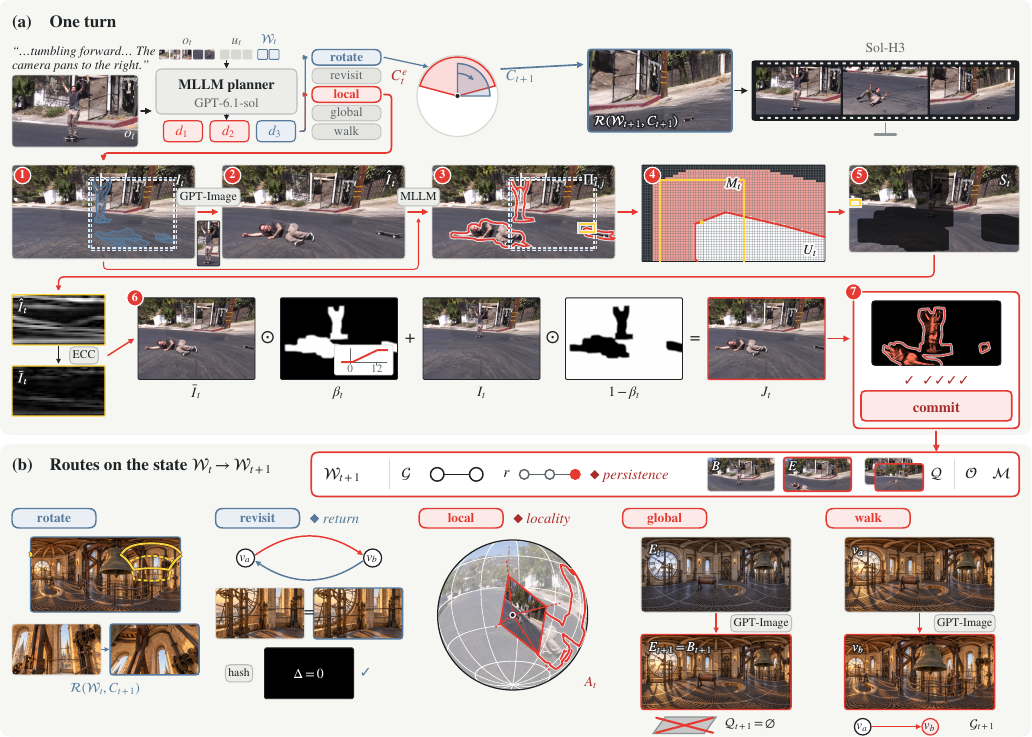}
\caption{\textbf{\sysname{} Overview.} Pretrained models read the explicit state, and only commits that pass the checks write it. (a)~One MBench-A turn with two local subject actions and a $45^\circ$ pan; the wedges show the editing camera (red) and the camera after the pan (blue). In steps 1--7, the MLLM marks the old and new footprints on the GPT-Image proposal, which are rasterized into $U_t$ and dilated into $M_t$; the proposal is registered on the stable background $S_t$ and blended with weight $\beta_t$; and the check finds no panorama change outside $A_t$. We recompute steps 4--6 from the records of this turn. (b)~The commit writes the state $\mathcal{W}_{t+1}$ (scene graph, revision history, and the base, composite and patches of this turn), and each route changes a different part of it; diamonds mark the parts that give return, locality and persistence. Rotation, revisit, global change and walking come from turns 3--6 of the session in \cref{fig:session}.}
\label{fig:method}
\end{figure}

\begin{figure}[tbp]
\centering
\includegraphics[width=\textwidth]{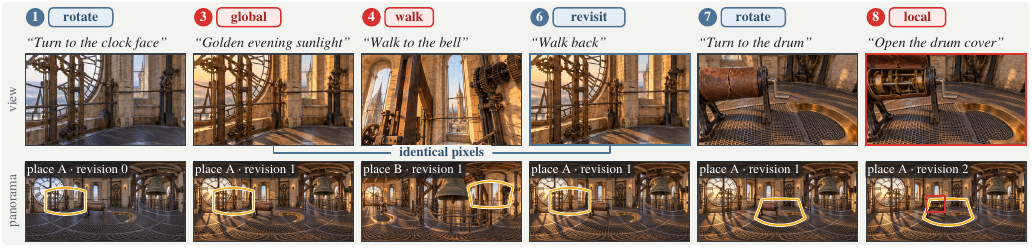}
\caption{\textbf{One session in one world.} The state carries every change: the light of turn~3 is still there after the walk to the bell and back, and turn~8 opens the drum under it. Turn~6 reproduces the view of turn~3 pixel for pixel, and turn~8 changes only the 2.1\% of the panorama pixels in the red box. Yellow outlines mark each camera; blue routes call no image generator. Turns~2 and~5 are not shown: turn~2 fails and keeps the state, and turn~5 tilts the camera.}
\label{fig:session}
\end{figure}

Where the world lives decides whether these properties hold. Video world models generate observations from camera moves, actions and event prompts, and streaming variants extend them to long, interactive rollouts~\cite{genie3_research2025,matrixgame3,hyworldplay,longlive}. Their world lives in the context of the generator, so a revisit synthesizes the view again and earlier edits can drift or disappear; after one camera round trip, five baselines, all video world models, end on a different scene (\cref{fig:teaser}, bottom row). 3D world models build an explicit scene, such as a mesh or a set of Gaussian surfels, that a renderer displays from new viewpoints and that supports regional edits~\cite{hunyuanworld,wonderworld,marble_official2025}. Among the systems with an explicit world state in \cref{tab:capabilities}, none documents all four of an MLLM planner, object edits, exact return and checks of an edit before it enters the state. World models for control, such as vision-language-action models~\cite{openvla,pi0} and joint-embedding predictive architectures~\cite{vjepa2}, output actions or embeddings and no editable images.

Multimodal large language models (MLLMs) and image generators such as Nano Banana and GPT-Image~\cite{nanobanana,gptimage,gptimage25} understand a scene, follow detailed instructions and accept both text and image inputs. We investigate whether these models can be turned into an interactive world model that users keep editing and exploring.

\sysname{} turns MLLMs into such a world model by keeping the world state outside every generator and letting the MLLMs write it (\cref{fig:method}). For each scene revision, the state stores a panorama and patches at native resolution; it also stores scene relations and facts observed in images. The MLLM GPT-6.1-sol~\cite{gpt61sol} maps each request and optional box to operations on this state, selects where a local edit is written, and checks every candidate before it becomes the next state. Rotations and revisits render the state and call no image generator; GPT-Image proposes candidate views, and the video generator Sol-H3~\cite{solh3} presents transitions without writing the state. Return, locality and persistence therefore follow from how each turn writes the state (\cref{sec:method_properties}), whereas a video world model has to reproduce them from its context. In one session of eight requests, a room keeps its new light after the user walks away and back, and a later edit builds on it (\cref{fig:session}).

A local edit is written inside a region, its \emph{support}, and \sysname{} selects this support after generation. An image generator can move a subject or change its silhouette beyond a region predicted in advance, so such a region can cut off part of the change (\cref{fig:support-ablation}). An MLLM therefore compares the source and the proposal and selects the old and new footprints and necessary effects.

\sysname{} achieves state-of-the-art scene consistency and camera return and the best CompBench editing scores. Against six world model baselines under the same MBench evaluator, it has the highest mean on all four environment-recovery metrics, with Epipolar 91.5 versus 0.1 to 74.0. It completes 85.3\% of camera round trips versus 0.4\% to 50.9\% (\cref{tab:main-results}). On CompBench, it has the best value in all 15 metric columns against 15 baselines (\cref{tab:compbench-sota}).

\paragraph{Contributions.}
\begin{itemize}
\item \textbf{The first interactive world model built on MLLMs.} \sysname{} composes navigation, object editing, environmental change and subject placement in one world built from a single image. Its state lives outside every generator, and view reuse, compositing inside the support and lookup by revision give return, locality and persistence by construction (\cref{tab:routes}, Propositions~1--3): all 46 revisits and all 97 local edits that pass the check satisfy them exactly, and over 21 sequential requests in one world the state keeps every earlier change (\cref{sec:exp_properties}).
\item \textbf{Local edits in the panorama state.} An MLLM compares the source with the proposal to select the old and new footprints of an object and their effects. On 112 local edits, completion rises from 34.8\% with support predicted before generation to 83.0\% with support selected after it (\cref{fig:ablation-judged}).
\item \textbf{State-of-the-art consistency and editing.} \sysname{} leads six baselines on MBench-A and 15 baselines on CompBench (\cref{sec:exp}).
\end{itemize}

\FloatBarrier
\section{Related Work}
\label{sec:related_work}
\subsection{World Representation and Memory}
Video world models maintain interaction by conditioning new frames on actions and visual context. Genie~3 supports navigation and promptable events~\cite{genie3_research2025}, and LongLive updates its context cache when the prompt changes~\cite{longlive}. Camera-aware context improves spatial continuity through memory retrieval in Matrix-Game~3.0~\cite{matrixgame3}, RGB-D generation in HunyuanWorld-Voyager~\cite{voyager}, and reframing of historical observations in HY-WorldPlay~\cite{hyworldplay}. WorldDirector adds planning of object and camera trajectories by a language model and preserves the appearance of dynamic objects that leave and return~\cite{worlddirector}. Genie~4 uses Gemini models to plan and verify steps on the frames it generates~\cite{genie4_official2026}.

Memory extends the visual context available to these generators. WorldMem retrieves frames by pose and time~\cite{worldmem}, Context-as-Memory by field-of-view overlap~\cite{contextasmemory}, and VMem through a surfel index~\cite{vmem}. Alaya-EVOKE maintains a geometry bank indexed by camera~\cite{alayaevoke}, while ActWorld stores event and object tokens with visual anchors in a memory of constant capacity~\cite{actworld}. Other approaches use spatial or planar memories or compress visual context~\cite{spmem,ppmem,worldtrace,worldt3}. Persistent Computational State saves and restores the generator's runtime state~\cite{pcs}. These memories condition a new synthesis on return.

Explicit 3D worlds support rendering and editing through scene representations. HunyuanWorld~1.0 reconstructs a mesh from a panoramic proxy~\cite{hunyuanworld}, and WonderWorld connects scenes through Gaussian surfels~\cite{wonderworld}. Marble edits and expands persistent navigable worlds by region, including weather, time of day and season~\cite{marble_official2025}. HOLODECK~2.0 and PERSIST update assets or latent 3D state~\cite{holodeck2,persist}. State updates from language also appear in WorldWeave, whose video synthesis reads an explicit 3D world~\cite{worldweave}, and StateFlow, which translates intent into transitions of persistent 3D state~\cite{stateflow}. WorldAgents adds new views from a generator to a reconstruction after checks by a vision-language model (VLM) in image and reconstruction space~\cite{worldagents}. \sysname{} uses a panorama and patches at native resolution as its editable visual state. Each physical change advances the revision, and a revisit selects a view of the current revision.

\subsection{MLLM Planning and Feedback}
Visual agents connect language planning, image generation and feedback. Yang et al.'s WorldGPT expands a request, edits a target image and animates the initial and final frames~\cite{worldgpt_yang2024}. GenArtist plans through tool trees~\cite{genartist}, and EditAgent routes steps to general and task-specific editors and then judges parallel candidates~\cite{editagent2026}. Agent Banana retains the editing paths that pass its checks, rolls back image history and edits layers in isolation~\cite{agentbanana}. IMAGAgent corrects multi-turn edits through feedback from several VLM experts~\cite{imagagent2026}. Ge et al.'s WorldGPT learns multimodal state transitions~\cite{worldgpt_ge2024}, and World Agent tests whether language models update an explicit world state from events in an ongoing narrative~\cite{worldagent2026}.

Agentic video systems connect this feedback to worlds that change over time. LingBot-World v2 uses a vision-language director for semantic rules, events and environmental changes, with SAM tracking objects for interaction~\cite{lingbotworldv2}. Regions that leave its context window tend to be regenerated on return. AgiWorld updates a world state from video chunks stored with geometry in spatial memory~\cite{agiworld2026}. ATAS adds clips to scene and event memories after review and state checks~\cite{atas2026}. \sysname{} gives MLLMs control over edits and navigation in a panorama state, and its transition videos read the endpoints of each turn.

\subsection{Image Editing and Consistency}
Image editors such as Nano Banana and GPT-Image follow complex instructions~\cite{nanobanana,gptimage,gptimage25}. Spherical editors extend this capability to panoramas. SE360 and World-Shaper edit spherical imagery~\cite{se360,worldshaper}, FocusDiff edits regions of indoor panoramas~\cite{focusdiff}, and Mover360 moves, inserts and removes objects~\cite{mover360}. ZONE extracts the region that an edit changes as a layer and blends it into the source image~\cite{zone}. \sysname{} selects the support from the source and the proposal to cover the old and new footprints of an object and their effects. It installs this region in a world that later operations navigate, revisit and edit again.

CompBench measures local, multi-object and implicit-reasoning edits~\cite{compbench}. Scene-memory benchmarks measure recovery after the camera leaves and returns~\cite{mbench,mind}, while WRBench tests whether events evolve outside the view~\cite{wrbench}. WorldFoundry provides infrastructure for inference and evaluation across video, 3D, 4D and interactive world systems~\cite{worldfoundry}. We evaluate both image editing and scene recovery. \Cref{tab:capabilities} compares world representations, MLLM planning, object edits, exact return and edit checks.

\begin{table}[tbp]
\centering\begingroup
\small
\setlength{\tabcolsep}{4pt}\renewcommand{\arraystretch}{1.2}
\newcommand{\capyes}{\tikz[baseline=-0.55ex]\fill[oeink] (0,0) circle (0.75ex);}%
\newcommand{\cappart}{\tikz[baseline=-0.55ex]{\fill[oeink] (0,-0.75ex) arc (-90:90:0.75ex) -- cycle;\draw[oeink,line width=0.45pt] (0,0) circle (0.75ex);}}%
\newcommand{\capno}{\tikz[baseline=-0.55ex]\draw[oeink,line width=0.45pt] (0,0) circle (0.75ex);}%
\caption{\textbf{Capability comparison.} \sysname{} combines an MLLM planner, object edits, exact return and edit checks in one persistent world state. Rows cover video world models with memory or agentic control, an agent that edits and animates a target image, and systems with an explicit world state; solid, half and empty circles mark capabilities that the sources document, document in part or do not document (\cref{sec:appendix_comparison_sources}).}
\label{tab:capabilities}
\begin{tabularx}{\linewidth}{l>{\raggedright\arraybackslash}X*{4}{>{\centering\arraybackslash}p{.09\linewidth}}}
\tabletoprule
\tablehead
System & World state & \shortstack{MLLM\\planner} & \shortstack{\smash[b]{Object}\\edits} & \shortstack{Exact\\return\vphantom{l}} & \shortstack{Edit\\checks} \\
\midrule
LingBot-World v2~\cite{lingbotworldv2} & Video context, key/value cache & \capyes & \cappart & \capno & \capno \\
WorldDirector~\cite{worlddirector} & Video context, memory of dynamic objects & \cappart & \capno & \cappart & \capno \\
Alaya-EVOKE~\cite{alayaevoke} & Geometry bank indexed by camera & \capno & \capno & \cappart & \capno \\
ActWorld~\cite{actworld} & Event and object tokens, visual anchors & \capno & \cappart & \cappart & \capno \\
LongLive~\cite{longlive} & Key/value recache, frames in memory & \capno & \capno & \capno & \capno \\
WorldGPT~\cite{worldgpt_yang2024} & Endpoint image after editing & \cappart & \capyes & \capno & \capno \\
Marble~\cite{marble_official2025} & Persistent splats or mesh & \capno & \capyes & \cappart & \capno \\
WorldWeave~\cite{worldweave} & Metric terrain, asset instances, interfaces & \capyes & \capno & \cappart & \capyes \\
StateFlow~\cite{stateflow} & Object geometry, poses, attributes & \capyes & \capyes & \cappart & \capno \\
WorldAgents~\cite{worldagents} & Views with poses after verification, 3D splats & \capyes & \capno & \capyes & \capyes \\
\midrule
\oursrow \sysname{} & Panorama, patches, scene graph, facts & \capyes & \capyes & \capyes & \capyes \\
\bottomrule
\end{tabularx}
\endgroup
\end{table}

\section{Method}
\label{sec:method}
\sysname{} stores each scene as a panorama with local patches and a revision history. A deterministic renderer produces its views (\cref{fig:method}). GPT-6.1-sol plans operations, selects the support of local edits, checks candidates and records observations. GPT-Image proposes endpoint images, and Sol-H3 generates transition videos. The executor applies each operation through one of five routes, and each route changes a fixed part of the state (\cref{tab:routes}). The state after each turn becomes the source of the next turn, and these update rules give return, locality and persistence.

\subsection{Problem Formulation}
\label{sec:method_overview}
A session is a sequence of requests $u_1,\dots,u_T$, each a free-form instruction with an optional box on the current view. A world model keeps a state $\mathcal W_t$ and a camera $C_t$, shows the view $o_t$, and answers each request with a new state, camera and view and a transition video,
\begin{equation}
(\mathcal W_{t+1},C_{t+1},o_{t+1})=F(\mathcal W_t,C_t,u_t),\qquad
V_t=\textsc{Present}(\mathcal W_t,C_t,\mathcal W_{t+1},C_{t+1},d_t),
\label{eq:turn}
\end{equation}
where $d_t$ is the sequence of operations from the planner. An interactive world requires return, locality and persistence (\cref{sec:intro}). In a video world model, one generator computes both $F$ and \textsc{Present}, so these properties depend on what it reproduces from its context.

\subsection{World State}
\label{sec:persistent_world}
\label{sec:persistent_world_state}
The world state is
\begin{equation}
\mathcal W_t=(\mathcal G_t,v_t,\{\mathcal P_{v,r}\},\mathcal O_t,\mathcal M_t),
\label{eq:persistent_world_state}
\end{equation}
where $\mathcal G_t$ is a graph of scenes and views with parent and portal relations, $v_t$ is the active scene, $\mathcal P_{v,r}$ is the visual state of scene $v$ at physical revision $r$, $\mathcal O_t$ holds object records with selections and affordances, and $\mathcal M_t$ holds facts observed in images. A visual state $\mathcal P_{v,r}=(B,E,\mathcal Q)$ stores two equirectangular images of $4096\times2048$ pixels around one camera center (a rendering base $B$ and a panorama $E$ into which local edits are composited) and a list $\mathcal Q=(Q_1,\dots,Q_K)$ of patches at native resolution in compositing order, each with a camera and an alpha map. The renderer $\mathcal R$ maps each pixel of camera $C$ to its ray $\mathbf d$ and composites
\begin{equation}
c_0=B\big(\pi(\mathbf d)\big),\qquad
c_k=\alpha_k(\mathbf d)\,Q_k(\mathbf d)+\big(1-\alpha_k(\mathbf d)\big)\,c_{k-1},
\label{eq:render}
\end{equation}
returning $c_K$ rounded to 8-bit color. Here $\pi(\mathbf d)=(\tfrac12+\tfrac{\lambda}{2\pi},\tfrac12-\tfrac{\varphi}{\pi})$ maps the longitude $\lambda$ and latitude $\varphi$ of $\mathbf d$ to panorama coordinates, sampled bilinearly, and $Q_k(\mathbf d)$ and $\alpha_k(\mathbf d)$ are bilinear samples of the color and alpha of patch $k$, with $\alpha_k=0$ outside it. The view is $o_t=\mathcal R(\mathcal P_{v_t,r_t},C_t)$ at the current revision $r_t$; $\mathcal R$ is deterministic and calls no model. A rotation changes $C_t$ and keeps $r$; a physical change advances $r$, so turning back shows the new revision.

Initialization outpaints the input image, registers the expansion and completes the panorama. Keeping the input image as a native patch reconstructs the initial view exactly in all 246 benchmark items that obtain a panorama (\cref{sec:appendix_prompt_contract}). \Cref{impl:sec:persistent_world_state} gives the projection and the initialization steps. The example worlds of \cref{fig:teaser,fig:session} start instead from a panorama that GPT-Image generates from text.

\subsection{Turn Transition}
\label{sec:routing_and_scope}
\label{sec:agentic_transition}
\label{sec:verified_commit}
Each visual request follows \cref{alg:turn}. A user describes a target in text or draws a box on the view; \sysname{} maps the box to a perspective crop and a spherical mask that wraps across the panorama border. The operation planner reads the request, the current view, the panorama context, the scene relations, facts from memory and the optional selection, and returns operations in order
\begin{equation}
d_t=(d_{t,1},\dots,d_{t,n_t}),\quad d_{t,i}=(z_{t,i},a_{t,i},\Omega_{t,i},q_{t,i}),
\label{eq:agent_plan}
\end{equation}
where $z_{t,i}$ is the route, $a_{t,i}$ the target, $\Omega_{t,i}$ the change that the operation may make and $q_{t,i}$ the executor arguments. \Cref{fig:method}b shows how each route changes the state, and \cref{tab:routes} gives the part of the visual state each one changes, the generator it calls and its check. Subject actions, including placement, follow the local or global route according to their scope. The executor composes the operations in order into one candidate endpoint. For example, it can apply a local edit at a fixed camera while the viewing camera pans. When a turn commits, it records the new state, camera, view and facts.

\begin{algorithm}[tbp]
\caption{\textbf{One visual turn.} A candidate writes the state only when no criterion fails and $\mathrm{Ready}_t$ holds, and presentation only reads the state.}
\label{alg:turn}
\small
\begin{algorithmic}[1]
\REQUIRE state $\mathcal W_t$, camera $C_t$, view $o_t$, request $u_t$
\STATE $d_t\gets\textsc{Plan}(u_t,o_t,\mathcal W_t,C_t)$\hfill$\triangleright$ ordered operations%
\IF{$d_t$ changes only the camera}
\STATE \rule{0pt}{2.7ex}$(\widehat{\mathcal W}_t,\hat C_t,\hat o_t)\gets\textsc{Render}(\mathcal W_t,C_t,d_t)$\hfill$\triangleright$ Eq.~\eqref{eq:render}%
\STATE $s\gets\textsc{Check}(o_t,\hat o_t,u_t)$\hfill$\triangleright$ MLLM%
\ELSE
\FOR{attempt $=1,2$}
\STATE $p\gets\textsc{EndpointPlan}(u_t,o_t,\mathcal W_t,d_t,\text{feedback})$
\STATE $(\widehat{\mathcal W}_t,\hat C_t,\hat o_t)\gets\textsc{Execute}(\mathcal W_t,C_t,d_t,p)$\hfill$\triangleright$ Eq.~\eqref{eq:scoped_update}%
\STATE $s\gets\textsc{Check}(o_t,\hat o_t,u_t)$\hfill$\triangleright$ MLLM%
\STATE \textbf{if} $\forall k\colon s_k\neq\textsc{Fail}$ \textbf{then break}
\STATE feedback $\gets$ failure record and proposal that failed
\ENDFOR
\ENDIF
\STATE $f_t\gets\textsc{Observe}(o_t,\hat o_t)$ if no $s_k=\textsc{Fail}$\hfill$\triangleright$ MLLM%
\IF{$\forall k\colon s_k\neq\textsc{Fail}$ \textbf{and} $\mathrm{Ready}_t$}
\STATE $(\mathcal W_{t+1},C_{t+1})\gets\textsc{Commit}(\widehat{\mathcal W}_t,\hat C_t,f_t)$
\ELSE
\STATE hold $(\mathcal W_t,C_t)$ and record the request, or stop
\ENDIF
\STATE $V_t\gets\textsc{Present}(\mathcal W_t,C_t,\mathcal W_{t+1},C_{t+1},d_t)$\hfill$\triangleright$ reads only%
\end{algorithmic}
\end{algorithm}

\begin{table}[!tbp]
\centering
\caption{\textbf{Transition routes.} Camera operations reuse the current visual revision. Local edits append a patch, global changes replace the panorama, and navigation starts a new scene. The right column lists the checks that each route requires.}
\label{tab:routes}
\includegraphics[width=\linewidth]{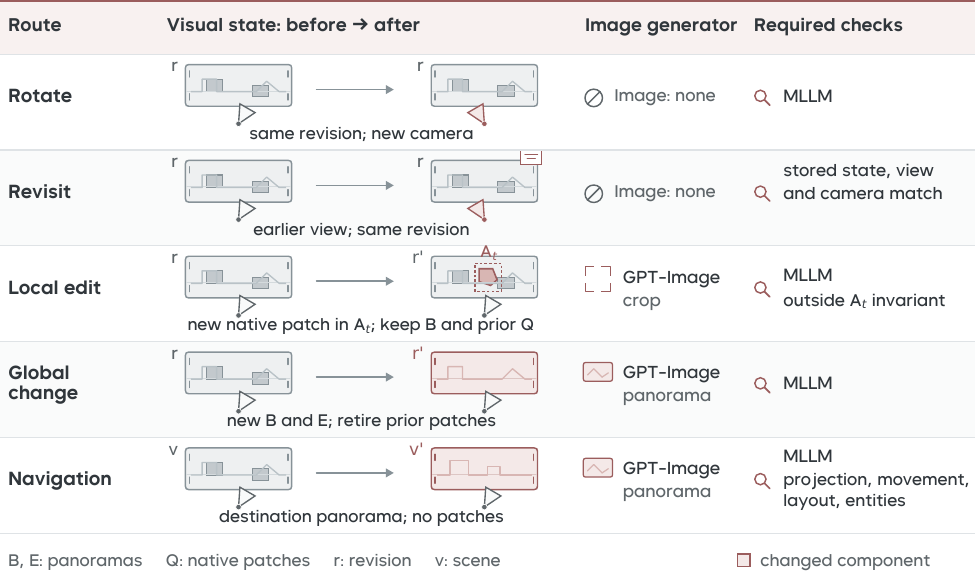}
\end{table}

An MLLM verifier compares the source view and the candidate $(\widehat{\mathcal W}_t,\hat C_t,\hat o_t)$ against the request on four criteria: the source description, the action, the change in the request and scene identity. For navigation it also checks projection, movement, layout and the continuity of entities. Each applicable criterion receives $s_k\in\{\textsc{Pass},\textsc{Uncertain},\textsc{Fail}\}$, and an observer records subject, property and value facts $f_t$ from both images. The state is updated by
\begin{equation}
(\mathcal W_{t+1},C_{t+1})=
\begin{cases}
\textsc{Commit}(\widehat{\mathcal W}_t,\hat C_t,f_t) & \text{if }\forall k\colon s_k\neq\textsc{Fail}\ \text{and}\ \mathrm{Ready}_t,\\
(\mathcal W_t,C_t) & \text{otherwise,}
\end{cases}
\label{eq:commit}
\end{equation}
where $\mathrm{Ready}_t$ requires a valid observation and the geometric and installation checks, such as \cref{eq:scoped_update}. An \textsc{Uncertain} criterion does not block a commit, and the commit keeps the \textsc{Uncertain} label. A candidate that fails is retried once on the same route, and the endpoint planner receives the failure record as feedback. A second failure keeps the current state and camera and records the request (a \emph{failure hold}), or ends the session. Each endpoint uses at most two image proposals and at most one Sol-H3 clip. \textsc{Commit} installs the panorama revision, camera, endpoint and facts as the source for the next request, so proposals that fail and video frames never enter memory.

\subsection{Rotation and Return}
\label{sec:deterministic_paths}
Rotation renders the current revision at a new orientation and calls no image or video generator. A revisit reuses a view in memory only when its physical revision, view identity, camera and rendering profile match. The executor checks that the hashes of the panorama state and arrival view agree (\cref{fig:method}b), and a return from another scene is also compared with a render at the target profile.

\subsection{Editing and Navigation}
\label{sec:transition_execution}
\begin{figure}[!tbp]
\centering
\includegraphics[width=\linewidth]{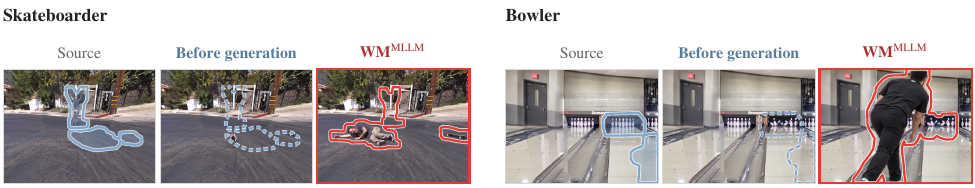}
\caption{\textbf{Support selection after generation.} The support that the MLLM selects after generation keeps the change in the request: the region predicted before generation cuts off the skateboarder and the bowler, and the support selected from the same proposal retains them.}
\label{fig:support-ablation}
\label{fig:support}
\end{figure}

A local edit or placement generates an endpoint crop $\hat I_t$ at the editing view $I_t$. The endpoint planner predicts a box and polygons inside it, but the generator can move an object, change its silhouette or add a shadow outside them (\cref{fig:support-ablation}). \sysname{} therefore selects the support from the proposal: a GPT-6.1-sol selector compares $I_t$ with $\hat I_t$ and returns polygons $\Pi_{t,1},\dots,\Pi_{t,L}$ that cover the old and new footprints and necessary effects such as shadows and contact.

Let $\bar I_t$ be the proposal after bounded rigid registration to the background of $I_t$. If registration fails its check, $\bar I_t=\hat I_t$. We blend
\begin{equation}
M_t=\delta_{16}\big(\textstyle\bigcup_j\Pi_{t,j}\big),\qquad
J_t=\beta_t\odot\bar I_t+(1-\beta_t)\odot I_t,
\label{eq:support}
\end{equation}
where $\delta_{16}$ rasterizes the polygons and dilates them by 16 pixels with a square filter, and the weight $\beta_t$ rises from 0 at the boundary of the mask $M_t$ to 1 at 12 pixels inside it, so $J_t=I_t$ outside $M_t$. An empty or invalid selection fails the operation without falling back to the planner's box, and an edit region that covers the whole view skips the selector. The compositor projects $M_t$ through the editing camera into the panorama as an alpha map $A_t$ and updates the panorama as
\begin{equation}
E_{t+1}=A_t\odot\widetilde E_t+(1-A_t)\odot E_t,\qquad
E_{t+1}[A_t{=}0]=E_t[A_t{=}0],
\label{eq:scoped_update}
\end{equation}
where $\widetilde E_t$ is the crop $J_t$ after reprojection. The executor aborts the operation if any pixel with $A_t=0$ changes (\cref{fig:method}a, step~7), and otherwise appends $J_t$ with alpha $M_t$ as a new native patch, which \cref{eq:render} composites over the base $B$.

A global change such as lighting, weather or season replaces both $B$ and $E$ and retires earlier patches, which would otherwise bring back the old appearance. Navigation to a new place generates a destination panorama for a new scene node.

\subsection{Video Presentation}
\label{sec:video_presentation}
Presentation never writes the state. Rotations and revisits around the same center render the panorama along a camera path between the two views with \cref{eq:render}. Other transitions, including returns from another scene, use Sol-H3 conditioned on the source and endpoint views and the request caption. SoL-Refiner~\cite{solrefiner} can optionally super-resolve the video. A failure hold repeats the current view. A view at rest can also play as a loop: Sol-H3 takes the same view as its first and last frame, with a caption that names motion in place. A loop writes nothing to the state, and no benchmark uses one.

\begin{samepage}
\subsection{Properties}
\label{sec:method_properties}
Three properties follow from \cref{tab:routes} and \cref{eq:commit}. A rotation or revisit reads its endpoint from the state, a local edit writes only inside its support, and each request starts from the preceding endpoint.

\noindent\textbf{Proposition 1 (Return).} A revisit that completes with the same physical revision, view identity, camera and rendering profile reproduces the earlier image pixel for pixel.
\emph{Proof.} The revisit reuses the image in the state after checking that the hashes of the panorama state and arrival view agree; a revisit from another scene is also compared with a deterministic render of the same state, camera and profile.

\noindent\textbf{Proposition 2 (Locality).} A local update keeps every panorama pixel with $A_t=0$. With $B$, the earlier patches, the camera and the profile fixed, it also keeps every pixel of the render whose ray has zero alpha in the new patch.
\emph{Proof.} \Cref{eq:scoped_update} copies $E_t$ where $A_t=0$, including its 8-bit values, and the executor aborts on any violation. For a ray with zero alpha, the renderer skips the new patch, so $c_{K+1}=c_K$.

\noindent\textbf{Proposition 3 (Persistence).} Each request starts from the endpoint of the preceding request, and a view lookup never installs a different physical revision.
\emph{Proof.} Retrieval and \textsc{Commit} require the source to equal the preceding endpoint, and the executor advances the source and visual state after each request. A hold keeps the visual state and camera, presentation never replaces the visual state, and the lookup rejects views of another revision.
\end{samepage}

\begin{figure}[!b]
\centering
\includegraphics[width=\textwidth]{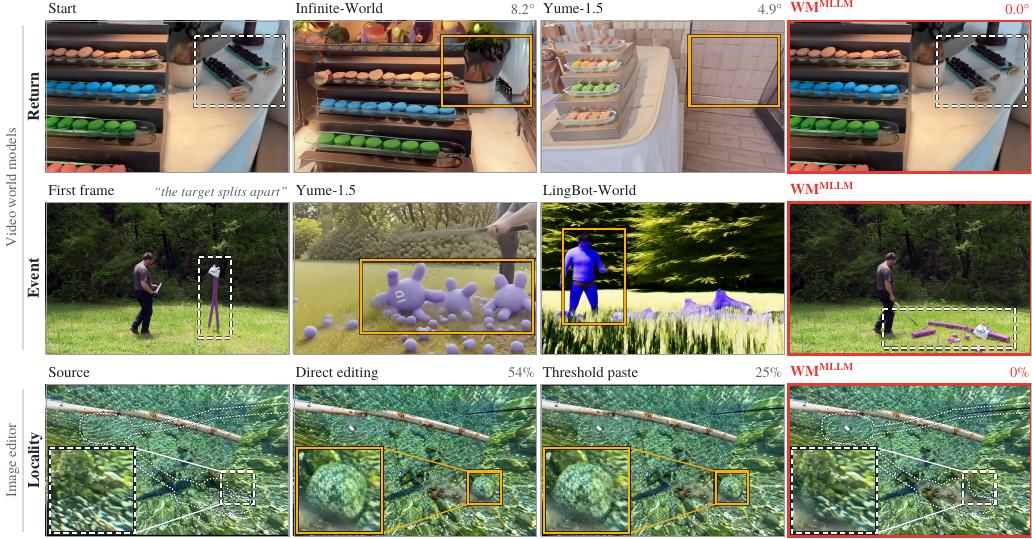}
\caption{\textbf{Where video world models and image editors fail.} \sysname{} returns to its starting view, keeps the pieces of the target after the camera round trip and confines local edits to their support. Angles give the distance of the final camera from its start, and percentages the fraction of pixels outside this support with a maximum channel change above 16/255; insets magnify the region in the box.}
\label{fig:paradigm-problems}
\label{fig:baseline-compare}
\end{figure}

\section{Experiments}
\label{sec:exp}
We use two public benchmarks, ablations on fixed image proposals and session studies. MBench-A measures environment memory~\cite{mbench}; 224 of our 227 environment items have a video from all six baselines, public world models conditioned on actions. CompBench~\cite{compbench} measures image editing on 2,709 GPT-Image proposals across local, multi-object and implicit-reasoning edits. We score all methods with the same public MBench evaluator, which preserves the ranking of the six baselines on 30 items per baseline. The benchmark run contains 158 WBench~\cite{wbenchpaper} navigation items and 100 MBench-A causal items, with 817 segments. Two paired studies with sessions of eight turns examine observation memory, the planner, the source of the next state and the endpoint check. A blind Claude Opus 5.5 judge assesses camera motion, edits and session outcomes from frame samples. \Cref{sec:appendix_eval_contract,sec:appendix_b1} give the protocols.

\subsection{MBench-A Memory and Camera Return}
\label{sec:exp_recovery}
We first test scene recovery after the camera leaves and returns (\cref{fig:paradigm-problems}, top row): \sysname{} has the highest mean on all four environment-recovery metrics against the six baselines (\cref{tab:main-results}). Its Epipolar mean is 91.5, 95\% confidence interval $[89.6,93.1]$, against 0.1 to 74.0, and on all four metrics its interval does not overlap that of any baseline. This result comes from reusing the state exactly: of the 454 segments in the 227 environment items, 88.3\% are geometric (215 rotations, 186 scene restores). On the 191 items whose \sysname{} segments contain no hold or terminal failure, Epipolar is still 93.9 versus at most 74.5, so holds do not explain the lead.

\begin{table}[tbp]
\centering
\fontsize{9}{10.3}\selectfont
\setlength{\tabcolsep}{3pt}\renewcommand{\arraystretch}{1.14}
\caption{\textbf{MBench-A results.} On MBench-A, \sysname{} leads all six baselines in environment recovery, pose return, usability and joint success; one baseline has higher camera motion. (a)~Environment-recovery means on 224 common items with one evaluator version; (b)~pose return on those items and blind frame judgments on 40 of them. Brackets give 95\% bootstrap CIs: over valid items for (a) and pose return, and over the 40 items, stratified by condition, for the frame judgments.}
\label{tab:main-results}
\textbf{(a) Environment recovery with one evaluator version (224 common items)}\par\smallskip
\setlength{\tabcolsep}{2.2pt}\fontsize{8.5}{9.8}\selectfont
\begin{tabularx}{\linewidth}{L{.22\linewidth}*{4}{R}}
\tabletoprule
\tablehead
Method & Epipolar $\uparrow$ & Reprojection $\uparrow$ & Lighting $\uparrow$ & Style $\uparrow$ \\
\midrule
\multicolumn{5}{l}{\emph{Baselines: public models conditioned on actions}} \\
HY-WorldPlay~\cite{hyworldplay} & \second{74.0} {\color{oemuted}\scriptsize [71.5, 76.3]} & \second{75.6} {\color{oemuted}\scriptsize [73.1, 78.0]} & \second{98.6} {\color{oemuted}\scriptsize [98.4, 98.7]} & \second{97.9} {\color{oemuted}\scriptsize [97.8, 98.1]} \\
Infinite-World~\cite{infiniteworld} & 59.5 {\color{oemuted}\scriptsize [56.5, 62.3]} & 60.9 {\color{oemuted}\scriptsize [58.1, 63.6]} & 94.8 {\color{oemuted}\scriptsize [94.5, 95.1]} & 96.2 {\color{oemuted}\scriptsize [96.0, 96.4]} \\
Matrix-Game 3.0~\cite{matrixgame3} & 47.5 {\color{oemuted}\scriptsize [43.8, 51.1]} & 53.8 {\color{oemuted}\scriptsize [50.7, 56.9]} & 94.4 {\color{oemuted}\scriptsize [93.9, 94.8]} & 93.5 {\color{oemuted}\scriptsize [93.1, 93.9]} \\
Yume-1.5~\cite{yume15} & 18.9 {\color{oemuted}\scriptsize [15.7, 22.3]} & 21.3 {\color{oemuted}\scriptsize [18.0, 24.7]} & 90.5 {\color{oemuted}\scriptsize [90.1, 91.0]} & 91.5 {\color{oemuted}\scriptsize [90.9, 92.0]} \\
Matrix-Game 2.0~\cite{matrixgame2} & 0.7 {\color{oemuted}\scriptsize [0.0, 1.6]} & 1.3 {\color{oemuted}\scriptsize [0.3, 2.5]} & 85.6 {\color{oemuted}\scriptsize [84.7, 86.5]} & 74.1 {\color{oemuted}\scriptsize [72.4, 75.8]} \\
LingBot-World~\cite{lingbotworld} & 0.1 {\color{oemuted}\scriptsize [0.0, 0.2]} & 2.1 {\color{oemuted}\scriptsize [1.2, 3.2]} & 89.6 {\color{oemuted}\scriptsize [89.0, 90.2]} & 85.1 {\color{oemuted}\scriptsize [84.4, 85.7]} \\
\midrule
\oursrow \sysname{} & \best{91.5} {\color{oemuted}\scriptsize [89.6, 93.1]} & \best{92.2} {\color{oemuted}\scriptsize [90.3, 93.8]} & \best{99.4} {\color{oemuted}\scriptsize [99.1, 99.6]} & \best{99.3} {\color{oemuted}\scriptsize [99.0, 99.5]} \\
\bottomrule
\end{tabularx}
\setlength{\tabcolsep}{3pt}\fontsize{9}{10.3}\selectfont

\par\vspace{6pt}
\textbf{(b) Camera behavior and frame quality [\%]}\par\smallskip
\begin{tabularx}{\linewidth}{L{.175\linewidth}XXXX}
\tabletoprule
\tablehead
Method & Pose return $\uparrow$ & Camera motion $\uparrow$ & Usable $\uparrow$ & Joint success $\uparrow$ \\
\midrule
\multicolumn{5}{l}{\emph{Baselines: public models conditioned on actions}} \\
HY-WorldPlay & \second{50.9} {\color{oemuted}\scriptsize [44.2, 57.6]} & \best{92.5} {\color{oemuted}\scriptsize [82.5, 100.0]} & \second{47.5} {\color{oemuted}\scriptsize [32.5, 62.5]} & \second{32.5} {\color{oemuted}\scriptsize [17.5, 47.5]} \\
Infinite-World & 25.0 {\color{oemuted}\scriptsize [19.2, 30.8]} & 42.5 {\color{oemuted}\scriptsize [27.5, 57.5]} & 37.5 {\color{oemuted}\scriptsize [22.5, 52.5]} & 2.5 {\color{oemuted}\scriptsize [0.0, 7.5]} \\
Matrix-Game 3.0 & 0.9 {\color{oemuted}\scriptsize [0.0, 2.2]} & 27.5 {\color{oemuted}\scriptsize [17.5, 37.5]} & 17.5 {\color{oemuted}\scriptsize [7.5, 27.5]} & 2.5 {\color{oemuted}\scriptsize [0.0, 7.5]} \\
Yume-1.5 & 20.5 {\color{oemuted}\scriptsize [15.6, 25.9]} & 15.0 {\color{oemuted}\scriptsize [5.0, 27.5]} & 15.0 {\color{oemuted}\scriptsize [5.0, 27.5]} & 0.0 {\color{oemuted}\scriptsize [0.0, 0.0]} \\
Matrix-Game 2.0 & 1.8 {\color{oemuted}\scriptsize [0.4, 3.6]} & 0.0 {\color{oemuted}\scriptsize [0.0, 0.0]} & 0.0 {\color{oemuted}\scriptsize [0.0, 0.0]} & 0.0 {\color{oemuted}\scriptsize [0.0, 0.0]} \\
LingBot-World & 0.4 {\color{oemuted}\scriptsize [0.0, 1.3]} & 0.0 {\color{oemuted}\scriptsize [0.0, 0.0]} & 0.0 {\color{oemuted}\scriptsize [0.0, 0.0]} & 0.0 {\color{oemuted}\scriptsize [0.0, 0.0]} \\
\midrule
\oursrow \sysname{} & \best{85.3} {\color{oemuted}\scriptsize [80.4, 89.7]} & \second{82.5} {\color{oemuted}\scriptsize [70.0, 92.5]} & \best{80.0} {\color{oemuted}\scriptsize [67.5, 92.5]} & \best{77.5} {\color{oemuted}\scriptsize [65.0, 90.0]} \\
\bottomrule
\end{tabularx}

\end{table}

\begin{figure*}[tbp]
\centering
\includegraphics[width=\textwidth]{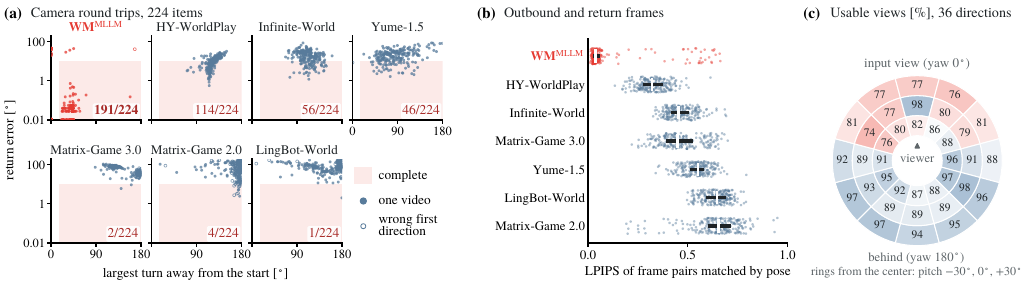}
\caption{\textbf{Return and consistency across models.} \sysname{} completes the most round trips and changes least between frames matched by pose. (a)~Camera round trips on the 224 common items, one point per video; the shaded region counts as complete (values below $0.01^\circ$ drawn at $0.01^\circ$). (b)~LPIPS between outbound and return frames matched by pose, per item (bars: quartiles and median). (c)~Share of views of 219 \sysname{} worlds that a blind model judge rates usable.}
\label{fig:consistency}
\end{figure*}

\sysname{} also completes the most camera round trips. A round trip is complete when the camera turns at least $15^\circ$ in the direction of the request during the first half and returns within $10^\circ$ of the start. \sysname{} completes 191 of the 224 items (85.3\%) and the six baselines 0.4\% to 50.9\% (\cref{tab:main-results}). \Cref{fig:consistency}a shows every item: \sysname{} turns a median of $43^\circ$ and returns within $0.1^\circ$ on 189 of the 224 items, while the baselines turn a median of $82^\circ$ to $173^\circ$ and complete 1 to 114 items. With outbound thresholds of $10^\circ$ to $20^\circ$ and return thresholds of $5^\circ$ to $15^\circ$, \sysname{} completes the same 191 items and the best baseline at most 158 (\cref{sec:appendix_recovery}). On 40 of these items, the judge labels from frame samples whether the camera goes out and back and whether the result is usable; joint success requires both and a final view that shows the starting content. \sysname{} reaches joint success on 31 of 40 videos and the baselines on 0 to 13 (\cref{tab:main-results}b).

A camera that returns does not guarantee a scene that returns. In the top row of \cref{fig:paradigm-problems}, Infinite-World and Yume-1.5 bring the camera back to about $8.2^\circ$ and $4.9^\circ$ from the start and still show a different counter. In the middle row, an event occurs during the round trip: Yume-1.5 and LingBot-World end on a different scene, and \sysname{} on its starting scene with the pieces of the target. Between outbound and return frames matched by pose, the median LPIPS is 0.037 for \sysname{} and 0.32 to 0.65 for the six baselines (\cref{fig:consistency}b). Over 36 directions, the judge rates 87.6\% $[85.7,89.4]$ of the views of \sysname{} usable, and most defects lie $30^\circ$ to $60^\circ$ from the input view, around its border (\cref{fig:consistency}c).

\subsection{Image Editing}
\label{sec:exp_compbench}
\sysname{} has the best value in all 15 CompBench columns, five metrics on three editing tasks, against 15 baselines in 16 configurations (\cref{tab:compbench-sota}). The baseline scores are those that CompBench reports, with HIVE in two configurations. LC-T is the local CLIP text score $\times100$, and LC-I is the foreground CLIP similarity to the target. PSNR, SSIM and LPIPS measure background preservation. The LC-I of \sysname{} is 0.845, 0.911 and 0.880 on the local, multi-object and implicit-reasoning groups, against 0.838, 0.883 and 0.874 for BAGEL, the strongest baseline on this measure. Because edits stay inside their support, background preservation shows the largest margins: PSNR is 30.440, 29.104 and 32.868 against at most 27.692, 26.849 and 28.891 for the baselines. \Cref{tab:compbench-sota} lists five of the baselines, including the strongest one of every column, and direct editing, which keeps the whole GPT-Image proposal.

\begin{table}[tbp]
\centering\small
\setlength{\tabcolsep}{7pt}\renewcommand{\arraystretch}{1.12}
\caption{\textbf{CompBench results.} \sysname{} has the best value in all 15 metric columns against 15 baselines. The rows include the strongest baseline of every column and direct editing, which keeps the whole GPT-Image proposal. Each panel reports one editing task.}
\label{tab:compbench-sota}
\begin{tabularx}{\linewidth}{lRRRRR}
\tabletoprule
\tablehead
Method & LC-T $\uparrow$ & LC-I $\uparrow$ & PSNR $\uparrow$ & SSIM $\uparrow$ & LPIPS $\downarrow$ \\
\midrule
\multicolumn{6}{l}{\emph{Local editing ($n=2{,}465$)}} \\
GoT~\cite{got} & 20.268 & 0.807 & 24.675 & 0.889 & 0.067 \\
Step1X-Edit~\cite{step1xedit} & 20.495 & 0.817 & 23.372 & 0.882 & 0.078 \\
BAGEL~\cite{bagel} & 21.059 & \second{0.838} & \second{27.692} & 0.935 & \second{0.045} \\
FLUX.1 Kontext~\cite{fluxkontext} & 21.328 & 0.821 & 25.613 & \second{0.941} & 0.050 \\
Qwen-Image-Edit~\cite{qwenimage} & 21.522 & 0.828 & 24.968 & 0.892 & 0.072 \\
Direct editing & \second{21.729} & 0.832 & 26.760 & 0.898 & 0.064 \\
\oursrow \sysname{} & \best{21.910} & \best{0.845} & \best{30.440} & \best{0.964} & \best{0.025} \\
\midrule
\multicolumn{6}{l}{\emph{Multi-object editing ($n=144$)}} \\
GoT~\cite{got} & 20.225 & 0.827 & 22.486 & 0.842 & 0.108 \\
Step1X-Edit~\cite{step1xedit} & 20.459 & 0.861 & 23.782 & 0.886 & 0.078 \\
BAGEL~\cite{bagel} & 20.856 & \second{0.883} & \second{26.849} & 0.932 & 0.051 \\
FLUX.1 Kontext~\cite{fluxkontext} & \second{21.020} & 0.868 & 26.278 & \second{0.952} & \second{0.048} \\
Qwen-Image-Edit~\cite{qwenimage} & 20.722 & 0.864 & 23.492 & 0.826 & 0.103 \\
Direct editing & 20.359 & 0.871 & 26.144 & 0.888 & 0.073 \\
\oursrow \sysname{} & \best{21.070} & \best{0.911} & \best{29.104} & \best{0.956} & \best{0.032} \\
\midrule
\multicolumn{6}{l}{\emph{Implicit reasoning ($n=100$)}} \\
GoT~\cite{got} & 19.246 & 0.821 & 24.889 & 0.860 & 0.088 \\
Step1X-Edit~\cite{step1xedit} & 19.175 & 0.847 & 23.408 & 0.869 & 0.083 \\
BAGEL~\cite{bagel} & 19.719 & \second{0.874} & \second{28.891} & 0.919 & \second{0.052} \\
FLUX.1 Kontext~\cite{fluxkontext} & 19.596 & 0.867 & 25.401 & \second{0.932} & 0.061 \\
Qwen-Image-Edit~\cite{qwenimage} & 20.046 & 0.859 & 22.815 & 0.775 & 0.124 \\
Direct editing & \second{20.193} & 0.863 & 27.624 & 0.893 & 0.063 \\
\oursrow \sysname{} & \best{20.241} & \best{0.880} & \best{32.868} & \best{0.970} & \best{0.019} \\
\bottomrule
\end{tabularx}
\end{table}

\subsection{Ablations}
\label{sec:exp_ablations}
\begin{figure}[tbp]
\centering
\includegraphics[width=\linewidth]{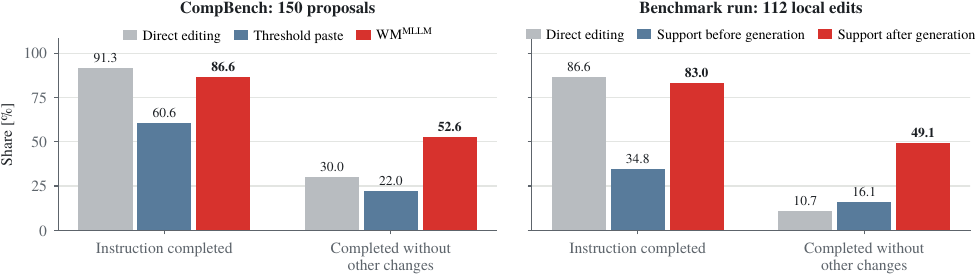}
\caption{\textbf{Blind evaluation of composition.} Keeping the whole proposal completes the most instructions, and support selected after generation most often completes the edit without other changes. Left: 150 CompBench proposals, with groups and subtasks weighted equally. Right: the first proposals of the 112 local edits of the benchmark run. Completion without other changes is computed after judging.}
\label{fig:ablation-judged}
\end{figure}

\paragraph{Applying a proposal.} \Cref{tab:compbench-control} keeps the 2,709 CompBench proposals fixed and changes only how a proposal enters the image: direct editing applies the whole proposal, threshold paste copies every pixel whose color change exceeds a threshold, and \sysname{} composites inside its support. In the bottom row of \cref{fig:paradigm-problems}, the whole GPT-Image proposal also adds a rock that the request does not ask for; direct editing and threshold paste change 54\% and 25\% of the pixels outside the support by more than 16 intensity levels, and \sysname{} none. \sysname{} has the best LC-I, SSIM and LPIPS in all three groups, and the 95\% confidence intervals of its LC-I differences against both alternatives exclude zero (\cref{sec:appendix_compbench}); threshold paste has the higher multi-object LC-T. In a blind comparison on 150 of these proposals, \sysname{} completes the request without any other change in 52.6\% of the cases, against 30.0\% for direct editing and 22.0\% for threshold paste (groups and subtasks weighted equally); direct editing completes the most instructions (\cref{fig:ablation-judged}).

\begin{table}[tbp]
\centering
\small
\setlength{\tabcolsep}{4pt}\renewcommand{\arraystretch}{1.12}
\caption{\textbf{Compositing the same proposals.} Compositing inside the support gives the best LC-I, SSIM and LPIPS in every group. All three rules use the same 2,709 GPT-Image proposals and evaluator.}
\label{tab:compbench-control}
\begin{tabularx}{\linewidth}{lRRRRR}
\tabletoprule
\tablehead
Composition & PSNR $\uparrow$ & SSIM $\uparrow$ & LPIPS $\downarrow$ & LC-T $\uparrow$ & LC-I $\uparrow$ \\
\midrule
\multicolumn{6}{l}{\emph{Local ($n=2{,}465$)}} \\
Direct editing & 26.760 & 0.898 & 0.064 & 21.729 & 0.832 \\
Threshold paste & 29.747 & 0.950 & 0.033 & 21.867 & 0.828 \\
\oursrow \sysname{} & \best{30.440} & \best{0.964} & \best{0.025} & \best{21.910} & \best{0.845} \\
\midrule
\multicolumn{6}{l}{\emph{Multi-object ($n=144$)}} \\
Direct editing & 26.144 & 0.888 & 0.073 & 20.359 & 0.871 \\
Threshold paste & 29.055 & 0.941 & 0.041 & \best{21.335} & 0.898 \\
\oursrow \sysname{} & \best{29.104} & \best{0.956} & \best{0.032} & 21.070 & \best{0.911} \\
\midrule
\multicolumn{6}{l}{\emph{Implicit reasoning ($n=100$)}} \\
Direct editing & 27.624 & 0.893 & 0.063 & 20.193 & 0.863 \\
Threshold paste & 31.867 & 0.954 & 0.029 & 20.014 & 0.851 \\
\oursrow \sysname{} & \best{32.868} & \best{0.970} & \best{0.019} & \best{20.241} & \best{0.880} \\
\bottomrule
\end{tabularx}
\end{table}

\paragraph{Selecting the support after generation.} On the benchmark run, we compare, for each local edit, the polygons that the endpoint planner predicts before generation with those that the selector returns after it. With the same registration, dilation and feathering on the first proposals of the 112 local edits, completion in blind evaluation rises from 34.8\% with the support before generation to 83.0\% after it, a difference of 48.2 points $[37.9,58.3]$ (\cref{fig:ablation-judged,fig:support-ablation}). The support selected after generation keeps the parts of the change that the region from the planner cuts off (\cref{sec:appendix_support_timing}).

\paragraph{Lookup by revision.} Without the revision in the view lookup, 60 of the 198 lookups of the benchmark run (30.3\%) would return a view from before a later edit, and each of them passes the hash checks (\cref{sec:appendix_revision_lookup}).

\subsection{Verification of the State Properties}
\label{sec:exp_properties}
\begin{figure}[tbp]
\centering
\includegraphics[width=\linewidth]{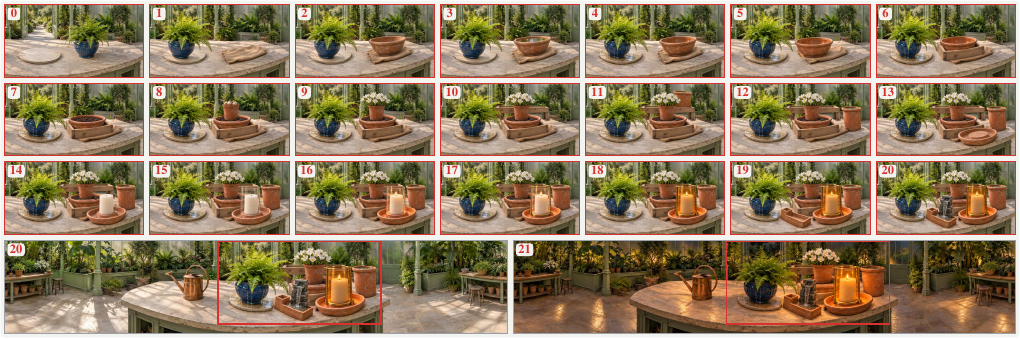}
\caption{\textbf{Playing in one world.} The state keeps every earlier change over 21 sequential requests. Rows 1--3: the red box at the start (0) and after each of requests 1--20, which change only this box (10.7\% of the panorama). Row 4: the panorama after request 20 and after request 21, which relights the whole panorama. \Cref{tab:session-requests} lists the requests.}
\label{fig:play}
\end{figure}

The benchmark run meets the properties of \cref{sec:method_properties} in every case. All 46 revisits match their panorama state and arrival view hashes, and all 195 geometric segments run without a generator. Of the 112 local edits, 97 pass the verifier, and all 97 leave every panorama pixel outside their support unchanged (\cref{eq:scoped_update}); in 92, the judge finds that the region matches the request (\cref{sec:appendix_locality}). Over 21 sequential requests in one world, the state keeps every earlier change, and requests 1--20 change only 10.7\% of the panorama (\cref{fig:play}). \Cref{fig:sessions-six} shows five more worlds with 20 or 21 sequential requests each. An agent writes each request of these sessions from the current state, and candidates that fail the checks are generated again, so these sessions do not measure success at the first attempt.

\subsection{Session Studies}
\label{sec:exp_sessions}
\paragraph{Observation memory.} Edits survive a return equally often without the observation memory. In 20 paired eight-turn sessions without video, the full system and the variant without the observer's facts and the retrieval of earlier images both keep the edit on 32 of 40 returns and complete 7 of 20 sessions. The panorama state carries the edits across returns in both settings (\cref{sec:appendix_session_counts}).

\paragraph{Planning, next state and endpoint checks.} Twenty new scripts compare the full system with a variant whose planner has a minimal prompt and with a variant that takes the last video frame as the next state. The primary results are close and their difference intervals include zero (\cref{tab:b1-outcomes}). A smaller study changes both the endpoint check and the proposal cap; on the same five sessions, both settings complete 23 of 40 turns. \Cref{sec:appendix_b1} describes these settings and the secondary reruns.

\subsection{Cost and Latency}
\label{sec:exp_cost}
The median time for state processing is 120\,s for rotation, 215\,s for navigation and 267\,s for multiple operations. Sol-H3 generates a 124-frame clip in a median of 71.9\,s on four H100 GPUs. In the multi-turn component study (\cref{sec:appendix_b1}), with two A6000 GPUs per job, the full system takes a median of 609.4\,s from request to first playback. Once a state is committed, every view of it renders from \cref{eq:render} without a model call. \Cref{sec:appendix_cost_contract} defines these measurements.

\section{Limitations}
\label{sec:limitations}
\begin{figure}[tbp]
\centering
\includegraphics[width=\textwidth]{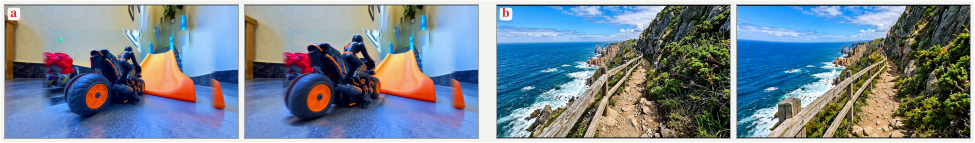}
\caption{\textbf{Failure cases.} Two endpoints from the benchmark run pass the verifier while the blind judge labels the change in the request absent. Each pair shows the source and the endpoint. (a)~For the request ``\ldots accelerates down a bright orange plastic slide'', the motorcycle stays on the floor beside the slide; both images use the same editing camera, and the verifier returns \textsc{Pass}. (b)~For the request ``The camera moves to the right.'', the coastal view shifts in the wrong direction; both images use the viewing camera, and the verifier returns \textsc{Uncertain}.}
\label{fig:failure-cases}
\end{figure}

The main limitation is that \sysname{} does not complete every request: 276 of the 280 failure holds of the benchmark run follow a rejection of the final endpoint, and these holds raise appearance scores while the motion in the request is missing. The endpoint verifier also passes some partial changes and rejects some complete ones: a blind judge labels as complete 190 of the 309 endpoints that pass the verifier and 83 of the 276 that it rejects (\cref{sec:appendix_judge}). Because each request starts from the endpoint of the preceding one, an error carries forward: the next segment completes in 100 of 153 cases after a correct endpoint commits and in 25 of 95 after an incorrect one. Weak camera motion and incorrect placement relations also cause failures. \Cref{fig:failure-cases} shows a motorcycle that stays beside the slide and a coastal view that shifts in the wrong direction.

The panorama represents a world around one camera center (\cref{sec:appendix_seams_media}). Translation and unseen regions require generation, and the generation log flags a seam in 208 of the 227 environment items. Video presentation also takes a median of about ten minutes from request to first playback (\cref{sec:exp_cost}).

The endpoint verifier and the blind judge lack human calibration, and frame samples do not establish the motion between two samples. The session studies have wide intervals and use new scripts on worlds from earlier studies.

\section{Conclusion}
\label{sec:conclusion}
\sysname{} turns MLLMs into a persistent, editable world model by keeping the world state outside every generator: the MLLMs write it under update rules that give return, locality and persistence, and generators only propose. Its panorama state carries object edits, environmental changes and navigation across turns, and over 21 sequential requests in one world it keeps every earlier change. Selecting the support after generation raises the completion of local edits from 34.8\% to 83.0\%. \sysname{} achieves the best mean on all four MBench-A recovery metrics against six baselines, completes the most camera round trips and has the best value in all 15 CompBench columns.

This design makes image generation part of an ongoing interaction in which users can revisit and revise the same world. More accurate endpoint decisions and faster video would extend the range and pace of these interactions.
\FloatBarrier

\label{page:main_end}
\section*{LLM Usage Statement}
In our system, GPT-6.1-sol~\cite{gpt61sol} plans operations, selects the support of local edits, checks endpoint candidates and records observations. GPT-Image~2~\cite{gptimage} supplies the CompBench proposals, GPT-Image~2.5~\cite{gptimage25} generates the endpoints in the other experiments, and GPT-Image also generates the initial panoramas. Sol-H3~\cite{solh3} generates the transition videos, and SoL-Refiner~\cite{solrefiner} optionally super-resolves them. Diagnostic images from earlier development also used Qwen-Image-Edit-2511. In the evaluation, Claude Opus 5.5 supplies all blind judgments and Qwen3-VL-8B-Instruct supplies the WBench~\cite{wbenchpaper} judge fields described in \cref{sec:appendix_scorers}; no human raters take part. Claude Code and OpenAI Codex serve as coding and writing assistants: they help to implement the system and the evaluation scripts, run the experiments, analyze the results, prepare the figures and tables, and draft and edit the text of this report. The authors direct and check this work and are responsible for the methods, results and final manuscript.

\phantomsection
\addcontentsline{toc}{section}{References}
\bibliographystyle{ieeetr}
\bibliography{references,bib/front,bib/method,bib/exp,bib/baselines}

\begin{thebibliography}{10}

\bibitem{worldfoundry}
J.~Tian, J.~Hou, B.~Li, X.~Lin, Z.~Jiao, W.~Zhang, Y.~Jin, T.~Nguyen, H.~Zheng,
  Z.~Zhang, J.~Yang, M.-H. Yang, F.~Heide, W.~Xiong, E.~Xie, and W.~Li,
  ``{WorldFoundry}: Towards unified infrastructure for world intelligence,''
  {\em arXiv preprint arXiv:2611.10771}, 2026.

\bibitem{genie3_research2025}
J.~Parker-Holder and S.~Fruchter, ``{Genie} 3: A new frontier for world
  models.''
  \url{https://deepmind.google/blog/genie-3-a-new-frontier-for-world-models/},
  Aug. 2025.
\newblock Accessed October 12, 2026.

\bibitem{matrixgame3}
Z.~Wang, Z.~Liu, J.~Li, K.~Huang, B.~Xu, F.~Kang, M.~An, P.~Wang, B.~Jiang,
  Y.~Wei, Y.~Xietian, J.~Pei, L.~Hu, B.~Jiang, H.~Xue, Z.~Wang, H.~Sun, W.~Li,
  W.~Ouyang, X.~He, Y.~Liu, Y.~Li, and Y.~Zhou, ``{Matrix-Game} 3.0: Real-time
  and streaming interactive world model with long-horizon memory,'' {\em arXiv
  preprint arXiv:2604.08995}, 2026.

\bibitem{hyworldplay}
W.~Sun, H.~Zhang, H.~Wang, J.~Wu, Z.~Wang, Z.~Wang, Y.~Wang, J.~Zhang, T.~Wang,
  and C.~Guo, ``{WorldPlay}: Towards long-term geometric consistency for
  real-time interactive world modeling,'' in {\em Proceedings of the
  International Conference on Machine Learning}, vol.~306 of {\em PMLR},
  pp.~117501--117523, 2026.

\bibitem{longlive}
S.~Yang, W.~Huang, R.~Chu, Y.~Xiao, Y.~Zhao, X.~Wang, M.~Li, E.~Xie, Y.~Chen,
  Y.~Lu, S.~Han, and Y.~Chen, ``{LongLive}: Real-time interactive long video
  generation,'' in {\em International Conference on Learning Representations},
  pp.~90037--90059, 2026.

\bibitem{hunyuanworld}
{HunyuanWorld Team}, ``{HunyuanWorld} 1.0: Generating immersive, explorable,
  and interactive {3D} worlds from words or pixels,'' {\em arXiv preprint
  arXiv:2507.21809}, 2025.

\bibitem{wonderworld}
H.-X. Yu, H.~Duan, C.~Herrmann, W.~T. Freeman, and J.~Wu, ``{WonderWorld}:
  Interactive {3D} scene generation from a single image,'' in {\em Proceedings
  of the IEEE/CVF Conference on Computer Vision and Pattern Recognition},
  pp.~5916--5926, 2025.

\bibitem{marble_official2025}
{World Labs}, ``{Marble}: A multimodal world model.''
  \url{https://www.worldlabs.ai/blog/marble-world-model}, Nov. 2025.
\newblock Editing documentation at docs.worldlabs.ai/marble/edit/pano-edit,
  accessed November 3, 2026.

\bibitem{openvla}
M.~J. Kim, K.~Pertsch, S.~Karamcheti, T.~Xiao, A.~Balakrishna, S.~Nair,
  R.~Rafailov, E.~P. Foster, P.~R. Sanketi, Q.~Vuong, T.~Kollar, B.~Burchfiel,
  R.~Tedrake, D.~Sadigh, S.~Levine, P.~Liang, and C.~Finn, ``{OpenVLA}: An
  open-source vision-language-action model,'' in {\em Proceedings of the 8th
  Conference on Robot Learning}, vol.~270 of {\em PMLR}, pp.~2679--2713, 2025.

\bibitem{pi0}
K.~Black, N.~Brown, D.~Driess, A.~Esmail, M.~R. Equi, C.~Finn, N.~Fusai,
  L.~Groom, K.~Hausman, B.~Ichter, S.~Jakubczak, T.~Jones, L.~Ke, S.~Levine,
  A.~Li-Bell, M.~Mothukuri, S.~Nair, K.~Pertsch, L.~X. Shi, L.~Smith,
  J.~Tanner, Q.~Vuong, A.~Walling, H.~Wang, and U.~Zhilinsky, ``$\pi_0$: A
  vision-language-action flow model for general robot control,'' in {\em
  Proceedings of Robotics: Science and Systems}, 2025.

\bibitem{vjepa2}
M.~Assran, A.~Bardes, D.~Fan, Q.~Garrido, R.~Howes, M.~Komeili, M.~Muckley,
  A.~Rizvi, C.~Roberts, K.~Sinha, A.~Zholus, S.~Arnaud, A.~Gejji, A.~Martin,
  F.~R. Hogan, D.~Dugas, P.~Bojanowski, V.~Khalidov, P.~Labatut, F.~Massa,
  M.~Szafraniec, K.~Krishnakumar, Y.~Li, X.~Ma, S.~Chandar, F.~Meier, Y.~LeCun,
  M.~Rabbat, and N.~Ballas, ``{V-JEPA} 2: Self-supervised video models enable
  understanding, prediction and planning,'' {\em arXiv preprint
  arXiv:2506.09985}, 2025.

\bibitem{nanobanana}
A.~Fortin, G.~Vernade, K.~Kampf, and A.~Reshi, ``Introducing {Gemini} 2.5
  {Flash} {Image}, our state-of-the-art image model.''
  \url{https://developers.googleblog.com/en/introducing-gemini-2-5-flash-image/},
  Aug. 2025.
\newblock Also known as Nano Banana; accessed October 25, 2026.

\bibitem{gptimage}
{OpenAI}, ``{ChatGPT Images 2.0} system card.''
  \url{https://deploymentsafety.openai.com/chatgpt-images-2-0}, Apr. 2026.
\newblock System card for the gpt-image-2 model family; accessed July 12, 2026.

\bibitem{gptimage25}
{OpenAI}, ``{ChatGPT Images 2.5} system card.''
  \url{https://deploymentsafety.openai.com/chatgpt-images-2-5}, Sept. 2026.
\newblock System card for GPT-Image-2.5-Sunburst and GPT-Image-2.5-Flare;
  accessed November 16, 2026.

\bibitem{gpt61sol}
{OpenAI}, ``Addendum to {GPT-6 Astra} system card: {GPT-6.1 Sol}.''
  \url{https://deploymentsafety.openai.com/gpt-6-1-sol}, Sept. 2026.
\newblock Accessed November 16, 2026.

\bibitem{solh3}
Y.~Li, J.~Yu, J.~Chen, H.~Li, S.~Xue, H.~Liu, P.~Luo, S.~Han, and E.~Xie,
  ``{Sol-H3}: Recursive self-improvement for {MiniMax-H3} inference
  acceleration on {Sol-Engine} across cloud and edge,'' {\em arXiv preprint
  arXiv:2609.35110}, 2026.

\bibitem{voyager}
T.~Huang, W.~Zheng, T.~Wang, Y.~Liu, Z.~Wang, J.~Wu, J.~Jiang, H.~Li, R.~W.~H.
  Lau, W.~Zuo, and C.~Guo, ``{Voyager}: Long-range and world-consistent video
  diffusion for explorable {3D} scene generation,'' {\em ACM Transactions on
  Graphics}, vol.~44, no.~6, pp.~1--15, 2025.
\newblock Article 245.

\bibitem{worlddirector}
H.~Wang, H.~Ouyang, Q.~Wang, W.~Wang, Q.~Bai, K.~L. Cheng, Y.~Yu, Y.~Li,
  Y.~Meng, Z.~Liu, Y.~Zeng, Y.~Shen, and Q.~Chen, ``{WorldDirector}: Building
  controllable world simulators with persistent dynamic memory,'' {\em arXiv
  preprint arXiv:2607.02517}, 2026.

\bibitem{genie4_official2026}
R.~Hadsell, D.~Rivas, and H.~Qi, ``{Genie} 4: A new generation of world
  simulation.''
  \url{https://deepmind.google/blog/genie-4-a-new-generation-of-world-simulation/},
  Oct. 2026.
\newblock Accessed November 3, 2026.

\bibitem{worldmem}
Z.~Xiao, Y.~Lan, Y.~Zhou, W.~Ouyang, S.~Yang, Y.~Zeng, and X.~Pan,
  ``{WorldMem}: Long-term consistent world simulation with memory,'' in {\em
  Advances in Neural Information Processing Systems}, vol.~38,
  pp.~49632--49652, 2025.

\bibitem{contextasmemory}
J.~Yu, J.~Bai, Y.~Qin, Q.~Liu, X.~Wang, P.~Wan, D.~Zhang, and X.~Liu,
  ``{Context as Memory}: Scene-consistent interactive long video generation
  with memory retrieval,'' in {\em Proceedings of the SIGGRAPH Asia 2025
  Conference Papers}, pp.~1--11, 2025.

\bibitem{vmem}
R.~Li, P.~Torr, A.~Vedaldi, and T.~Jakab, ``{VMem}: Consistent interactive
  video scene generation with surfel-indexed view memory,'' in {\em Proceedings
  of the IEEE/CVF International Conference on Computer Vision},
  pp.~25690--25699, 2025.

\bibitem{alayaevoke}
Y.~Yin, G.~Wang, Y.~Zhan, C.~Li, K.~Zhang, and F.~Zhao, ``{Alaya-EVOKE}: From
  linear-scaling supervision to endless world,'' {\em arXiv preprint
  arXiv:2608.13546}, 2026.

\bibitem{actworld}
Z.~Xiong, Y.~Song, H.~Kang, Q.~Yan, L.~Jiang, J.~Yang, Z.~Fu, S.~Fotiadis,
  A.~Wang, Z.~Liu, B.~Liu, Y.~Yang, X.~Lu, and N.~Jacobs, ``{ActWorld}: From
  explorable to interactive world model via action-aware memory,'' in {\em
  Advances in Neural Information Processing Systems}, 2026.

\bibitem{spmem}
T.~Wu, S.~Yang, R.~Po, Y.~Xu, Z.~Liu, D.~Lin, and G.~Wetzstein, ``Video world
  models with long-term spatial memory,'' in {\em Advances in Neural
  Information Processing Systems}, vol.~38, pp.~49371--49393, 2025.

\bibitem{ppmem}
Y.~He, B.~Tan, Z.~Gao, Y.~Shen, Y.-J. Liu, N.~Xue, and Y.~Xu, ``Persistent
  planar memory for video world models,'' in {\em Advances in Neural
  Information Processing Systems}, 2026.

\bibitem{worldtrace}
X.~Wu, S.~Elflein, J.~Lucas, O.~Russakovsky, L.~Leal-Taix{\'e}, D.~Paschalidou,
  J.~Lorraine, and A.~O{\v{s}}ep, ``Addressable memory for video world
  models,'' in {\em Advances in Neural Information Processing Systems}, 2026.

\bibitem{worldt3}
J.~Wu, Z.~Li, C.~Yang, C.~Guo, L.~Yang, Y.~Zhang, X.~Zhang, H.~Hu,
  B.~Rymkowski, S.~Sridhar, Y.~Yan, and X.~Huang, ``{World-T3}: Test-time
  training as persistent memory for real-time video world models,'' {\em arXiv
  preprint arXiv:2610.17434}, 2026.

\bibitem{pcs}
Z.~Lin, ``Persistent computational state: A session-centric runtime for
  generative world models,'' {\em arXiv preprint arXiv:2607.21686}, 2026.

\bibitem{holodeck2}
Z.~Bian, R.~Ren, Y.~Yang, and C.~Callison-Burch, ``{HOLODECK} 2.0:
  Vision-language-guided {3D} world generation with editing,'' {\em arXiv
  preprint arXiv:2508.05899}, 2025.

\bibitem{persist}
S.~Garcin, T.~Walker, S.~McDonagh, T.~Pearce, H.~Bilen, T.~He, K.~Wang, and
  J.~Bian, ``Beyond pixel histories: World models with persistent {3D} state,''
  in {\em Proceedings of the International Conference on Machine Learning},
  vol.~306 of {\em PMLR}, pp.~34048--34070, 2026.

\bibitem{worldweave}
Y.~Huang, L.~Jiang, Q.~Hao, C.~Chen, B.~Wu, X.~Ren, X.~He, and D.~Zhao,
  ``{WorldWeave}: Growing persistent geometric worlds for video generation,''
  {\em arXiv preprint arXiv:2609.34221}, 2026.

\bibitem{stateflow}
Y.~Yin, Z.~Li, L.~Deng, H.~Li, S.~Zhao, J.~Liu, W.~Huang, M.~Wang, T.~Fu,
  Y.~Wang, P.-S. Wang, X.~Jin, Y.~Zhao, and Y.~Wei, ``{StateFlow}: Building,
  evolving, and accessing {3D} world states for previsualization,'' {\em arXiv
  preprint arXiv:2608.12314}, 2026.

\bibitem{worldagents}
Z.~Erko{\c{c}}, A.~Dai, and M.~Nie{\ss}ner, ``{WorldAgents}: Can foundation
  image models be agents for {3D} world models?,'' in {\em Proceedings of the
  European Conference on Computer Vision}, vol.~17066 of {\em Lecture Notes in
  Computer Science}, pp.~474--491, 2026.

\bibitem{worldgpt_yang2024}
D.~Yang, L.~Hu, Y.~Tian, Z.~Li, C.~Kelly, B.~Yang, C.~Yang, and Y.~Zou,
  ``{WorldGPT}: A {Sora}-inspired video {AI} agent as rich world models from
  text and image inputs,'' {\em arXiv preprint arXiv:2403.07944}, 2024.

\bibitem{genartist}
Z.~Wang, A.~Li, Z.~Li, and X.~Liu, ``{GenArtist}: Multimodal {LLM} as an agent
  for unified image generation and editing,'' in {\em Advances in Neural
  Information Processing Systems}, vol.~37, pp.~128374--128395, 2024.

\bibitem{editagent2026}
H.~Xu, T.~Wang, J.~Li, J.~Mu, Y.~Xiong, J.~Zhang, Z.~Tu, and M.~Xu,
  ``{EditAgent}: Rethinking agentic image editing,'' {\em arXiv preprint
  arXiv:2610.39664}, 2026.

\bibitem{agentbanana}
R.~Ye, J.~Zhang, Z.~Liu, Z.~Zhu, S.~Yang, L.~Li, T.~Fu, F.~Dernoncourt,
  Y.~Zhao, J.~Zhu, R.~Rossi, W.~Chai, and Z.~Tu, ``{Agent Banana}:
  High-fidelity image editing with agentic thinking and tooling,'' {\em arXiv
  preprint arXiv:2602.09084}, 2026.

\bibitem{imagagent2026}
F.~Shen, C.~Xie, L.~Wang, Z.~Zhang, X.~Jiang, X.~Du, J.~Tang, and T.-S. Chua,
  ``{IMAGAgent}: Orchestrating multi-turn image editing via constraint-aware
  planning and reflection,'' {\em arXiv preprint arXiv:2603.29602}, 2026.

\bibitem{worldgpt_ge2024}
Z.~Ge, H.~Huang, M.~Zhou, J.~Li, G.~Wang, S.~Tang, and Y.~Zhuang, ``{WorldGPT}:
  Empowering {LLM} as multimodal world model,'' in {\em Proceedings of the 32nd
  ACM International Conference on Multimedia}, pp.~7346--7355, 2024.

\bibitem{worldagent2026}
W.~Chen, W.~Zhang, N.~An, Y.~Liu, and L.~Lin, ``{World Agent}: Can language
  models keep a world running?,'' {\em arXiv preprint arXiv:2609.32692}, 2026.

\bibitem{lingbotworldv2}
Z.~Gao, Q.~Wang, J.~Zhu, J.~Chen, Z.~Liu, Q.~Bai, J.~Wang, Y.~Yuan, H.~Wang,
  Y.~Lu, K.~L. Cheng, H.~Zhang, J.~Gao, T.~Feng, Y.~Liu, Y.~Yao, Y.~Xu, X.~Zhu,
  Y.~Shen, and H.~Ouyang, ``Infinite worlds with versatile interactions,'' {\em
  arXiv preprint arXiv:2607.07534}, 2026.

\bibitem{agiworld2026}
S.~Fang, Y.~Mei, H.~Jiang, X.~Sun, H.~Zheng, H.~Tan, K.~Sunkavalli,
  G.~Pavlakos, Q.~Huang, and Y.~Hu, ``{AgiWorld}: An agentic video world model
  for building persistent worlds,'' {\em arXiv preprint arXiv:2611.02341},
  2026.

\bibitem{atas2026}
W.~Wang, H.~Zhang, Y.~Li, A.~Liu, Z.~Zhang, H.~Zhao, Y.~He, D.~Y. Chen,
  F.~Chen, and B.~Zhuang, ``Any time, any space: A world-state-centric
  framework for agentic video generation,'' {\em arXiv preprint
  arXiv:2610.20571}, 2026.

\bibitem{se360}
H.~Zhong, F.-L. Zhang, A.~Chalmers, and T.~Rhee, ``{SE360}: Semantic edit in
  360{\textdegree} panoramas via hierarchical data construction,'' in {\em
  Proceedings of the AAAI Conference on Artificial Intelligence}, vol.~40,
  pp.~13512--13520, 2026.

\bibitem{worldshaper}
D.~Liang, Y.~Liu, J.~Jia, Y.~Zhao, and R.~W.~H. Lau, ``{World-Shaper}: A
  unified framework for 360{\textdegree} panoramic editing,'' in {\em
  Proceedings of the International Conference on Machine Learning}, vol.~306 of
  {\em PMLR}, pp.~73275--73300, 2026.

\bibitem{focusdiff}
D.-K. Vo, N.-T. Le-Hinh, V.-T. Huynh, T.~V. Nguyen, M.-T. Tran, and T.-N. Le,
  ``Toward 360-degree indoor panorama editing via tuning-free diffusion model
  with refocusing cross-attention,'' in {\em Computational Collective
  Intelligence}, vol.~16909 of {\em Lecture Notes in Computer Science},
  pp.~256--270, Springer Nature Switzerland, 2027.
\newblock ICCCI 2026; first online 21 September 2026.

\bibitem{mover360}
H.~Zhong, F.-L. Zhang, A.~Chalmers, and T.~Rhee, ``{Mover360}: Controllable
  object manipulation in 360$^{\circ}$ panoramic images,'' {\em arXiv preprint
  arXiv:2608.23238}, 2026.

\bibitem{zone}
S.~Li, B.~Zeng, Y.~Feng, S.~Gao, X.~Liu, J.~Liu, L.~Li, X.~Tang, Y.~Hu, J.~Liu,
  and B.~Zhang, ``{ZONE}: Zero-shot instruction-guided local editing,'' in {\em
  Proceedings of the IEEE/CVF Conference on Computer Vision and Pattern
  Recognition}, pp.~6254--6263, 2024.

\bibitem{compbench}
B.~Jia, W.~Huang, Y.~Tang, J.~Qiao, J.~Liao, S.~Cao, F.~Zhao, Z.~Feng, Z.~Gu,
  Z.~Yin, L.~Bai, W.~Ouyang, L.~Chen, F.~Zhao, Z.~Wang, Y.~Xie, and S.~Lin,
  ``{CompBench}: Benchmarking complex instruction-guided image editing,'' in
  {\em Proceedings of the IEEE/CVF Conference on Computer Vision and Pattern
  Recognition}, pp.~1112--1122, 2026.

\bibitem{mbench}
S.~Zhang, Z.~Zhang, S.~Huang, Z.~Tang, H.~Wang, C.~Dai, M.~Chen, Y.~Li, Y.~Li,
  Y.~Chen, H.~Liu, C.~Li, J.~Lyu, and Y.~Duan, ``{MBench}: A comprehensive
  benchmark on memory capability for video world models,'' {\em arXiv preprint
  arXiv:2606.00793}, 2026.

\bibitem{mind}
Y.~Ye, X.~Lu, Y.~Jiang, Y.~Gu, R.~Zhao, Q.~Liang, J.~Pan, F.~Zhang, W.~Wu, and
  A.~J. Wang, ``{MIND}: Benchmarking memory consistency and action control in
  world models,'' {\em arXiv preprint arXiv:2602.08025}, 2026.

\bibitem{wrbench}
J.~Lu, D.~Zhu, H.~Shi, L.~Cai, G.~Tang, Y.~Chen, J.~Cao, D.~Tang, Y.~Zhang,
  Y.~Dai, and X.~Ju, ``Current world models lack a persistent state core,''
  {\em arXiv preprint arXiv:2606.20545}, 2026.

\bibitem{solrefiner}
H.~Liu, T.~Ye, S.~Xue, S.~Gao, Y.~Li, J.~Chen, H.~Li, J.~Yu, D.~Wang, R.~Zhang,
  L.~Zhu, J.-W. Bian, S.~Han, and E.~Xie, ``{SoL-Refiner}: Speed-of-light
  one-step refinement for high-resolution video,'' {\em arXiv preprint
  arXiv:2609.37969}, 2026.

\bibitem{wbenchpaper}
K.~Ying, H.~Hu, S.~Ren, J.~Li, F.~Chen, Z.~Wang, X.~Cao, X.~Cai, and H.~Ding,
  ``{WBench}: A comprehensive multi-turn benchmark for interactive video world
  model evaluation,'' {\em arXiv preprint arXiv:2605.25874}, 2026.

\bibitem{infiniteworld}
R.~Wu, X.~He, M.~Cheng, T.~Yang, Y.~Zhang, Z.~Kang, X.~Cai, X.~Wei, C.-L. Guo,
  C.~Li, and M.-M. Cheng, ``{Infinite-World}: Scaling interactive world models
  to 1000-frame horizons via pose-free hierarchical memory,'' in {\em
  Proceedings of the International Conference on Machine Learning}, vol.~306 of
  {\em PMLR}, pp.~135994--136007, 2026.

\bibitem{yume15}
X.~Mao, Z.~Li, C.~Li, X.~Xu, K.~Ying, and K.~Zhang, ``{Yume1.5}: A
  text-controlled interactive world generation model,'' in {\em Proceedings of
  the IEEE/CVF Conference on Computer Vision and Pattern Recognition},
  pp.~7752--7761, 2026.

\bibitem{matrixgame2}
X.~He, C.~Peng, Z.~Liu, B.~Wang, Y.~Zhang, Q.~Cui, F.~Kang, B.~Jiang, M.~An,
  Y.~Ren, B.~Xu, H.-X. Guo, K.~Gong, S.~Wu, W.~Li, X.~Song, Y.~Liu, Y.~Li, and
  Y.~Zhou, ``{Matrix-Game} 2.0: An open-source, real-time, and streaming
  interactive world model,'' {\em arXiv preprint arXiv:2508.13009}, 2025.

\bibitem{lingbotworld}
{Robbyant Team}, Z.~Gao, Q.~Wang, Y.~Zeng, J.~Zhu, K.~L. Cheng, Y.~Li, H.~Wang,
  Y.~Xu, S.~Ma, Y.~Chen, J.~Liu, Y.~Cheng, Y.~Yao, J.~Zhu, Y.~Meng, K.~Zheng,
  Q.~Bai, J.~Chen, Z.~Shen, Y.~Yu, X.~Zhu, Y.~Shen, and H.~Ouyang, ``Advancing
  open-source world models,'' {\em arXiv preprint arXiv:2601.20540}, 2026.

\bibitem{got}
R.~Fang, C.~Duan, K.~Wang, L.~Huang, H.~Li, H.~Tian, S.~Yan, W.~Yu, X.~Zeng,
  J.~Dai, X.~Liu, and H.~Li, ``{GoT}: Unleashing reasoning capability of {MLLM}
  for visual generation and editing,'' in {\em Advances in Neural Information
  Processing Systems}, vol.~38, pp.~67680--67708, 2025.

\bibitem{step1xedit}
S.~Liu, Y.~Han, P.~Xing, F.~Yin, R.~Wang, W.~Cheng, J.~Liao, Y.~Wang, H.~Fu,
  C.~Han, G.~Li, Y.~Peng, Q.~Sun, J.~Wu, Y.~Cai, Z.~Ge, R.~Ming, L.~Xia,
  X.~Zeng, Y.~Zhu, B.~Jiao, X.~Zhang, G.~Yu, and D.~Jiang, ``{Step1X-Edit}: A
  practical framework for general image editing,'' {\em arXiv preprint
  arXiv:2504.17761}, 2025.

\bibitem{bagel}
C.~Deng, D.~Zhu, K.~Li, C.~Gou, F.~Li, Z.~Wang, S.~Zhong, W.~Yu, X.~Nie,
  Z.~Song, G.~Shi, and H.~Fan, ``Emerging properties in unified multimodal
  pretraining,'' {\em arXiv preprint arXiv:2505.14683}, 2025.

\bibitem{fluxkontext}
{Black Forest Labs}, S.~Batifol, A.~Blattmann, F.~Boesel, S.~Consul, C.~Diagne,
  T.~Dockhorn, J.~English, Z.~English, P.~Esser, S.~Kulal, K.~Lacey, Y.~Levi,
  C.~Li, D.~Lorenz, J.~M{\"u}ller, D.~Podell, R.~Rombach, H.~Saini, A.~Sauer,
  and L.~Smith, ``{FLUX.1 Kontext}: Flow matching for in-context image
  generation and editing in latent space,'' {\em arXiv preprint
  arXiv:2506.15742}, 2025.

\bibitem{qwenimage}
C.~Wu, J.~Li, J.~Zhou, J.~Lin, K.~Gao, K.~Yan, S.-m. Yin, S.~Bai, X.~Xu,
  Y.~Chen, Y.~Chen, Z.~Tang, Z.~Zhang, Z.~Wang, A.~Yang, B.~Yu, C.~Cheng,
  D.~Liu, D.~Li, H.~Zhang, H.~Meng, H.~Wei, J.~Ni, K.~Chen, K.~Cao, L.~Peng,
  L.~Qu, M.~Wu, P.~Wang, S.~Yu, T.~Wen, W.~Feng, X.~Xu, Y.~Wang, Y.~Zhang,
  Y.~Zhu, Y.~Wu, Y.~Cai, and Z.~Liu, ``{Qwen-Image} technical report,'' {\em
  arXiv preprint arXiv:2508.02324}, 2025.

\end{thebibliography}
\clearpage
\appendix
\crefalias{section}{appendix}\crefalias{subsection}{subappendix}\crefalias{subsubsection}{subsubappendix}
\setlength{\parskip}{2pt plus .5pt}
\renewcommand{\thesection}{A\arabic{section}}
\setcounter{figure}{0}
\setcounter{table}{0}
\renewcommand{\thefigure}{A\arabic{figure}}
\renewcommand{\thetable}{A\arabic{table}}
\section{Implementation Details}
\label{impl:sec:method}
\subsection{Panorama State}
\label{impl:sec:persistent_world_state}
The equirectangular panorama (ERP) has resolution $4096\times2048$. Each perspective pixel samples a ray through its center. For a view of width $w$ and height $h$, the camera ray at pixel $(i,j)$ is proportional to
\begin{equation}
\left(\left(2\frac{i+1/2}{w}-1\right)\tan\frac\theta2,
\left(1-2\frac{j+1/2}{h}\right)\frac hw\tan\frac\theta2,1\right),
\label{impl:eq:camera_rays}
\end{equation}
where $i=0,\ldots,w-1$, $j=0,\ldots,h-1$ and the default horizontal field of view is $\theta=90^\circ$. Yaw and pitch rotate the rays into world directions. The projection $\pi$ in \cref{eq:render} maps their longitude and latitude to ERP coordinates, wrapping horizontally and clamping vertically. The same rays sample patch colors and alpha bilinearly, with later patches composited last.

\begin{figure}[!hb]
\centering
\includegraphics[width=\linewidth]{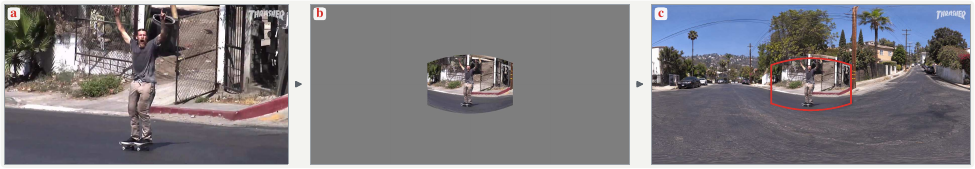}
\caption{\textbf{Initialization from one image.} The input stays in the state as a native patch, so the render at the initial camera reproduces it exactly. (a)~Input view of the MBench-A item of \cref{fig:method}. (b)~The input after perspective expansion, projected into the panorama frame; it covers 8.9\% of the panorama. (c)~The panorama after completion and registration to the input; the red outline marks the native patch.}
\label{fig:init}
\end{figure}

Initialization removes screen padding, registers the perspective expansion and the ERP completion to the input image, and preserves the input as a native patch, so the render at the initial camera reproduces the input exactly (\cref{fig:init}). A 64-pixel collar outside the source patch corrects residual color differences. A seam check allows one retry and at most two local repairs.

\subsection{Updates and Memory}
\label{impl:sec:routing_and_scope}
\label{impl:sec:transition_execution}
A selection in the application binds a screen box, the camera and the scene revision to a perspective crop. The operation planner returns the targets, scope, cameras and endpoint instructions. For a local edit, the endpoint planner predicts a box and polygons; the first polygon identifies the object. After generation, the selector marks the old and new footprints and their effects, including changes beyond the planner's box. The support that the selector marks is dilated by 16 pixels and feathered 12 pixels inward. Composition copies the source elsewhere and aborts if any ERP pixel changes outside the support.

Physical changes advance the scene revision (\cref{fig:session,fig:play}); rotations keep it. A revisit around the same center renders the state in memory under the conditions in \cref{sec:deterministic_paths}. The Back command activates the current state of the parent scene.

\label{impl:sec:verified_commit}
The endpoint verifier checks the source description, the action, the change in the request and scene identity. Checking precedes observation and installation. A candidate that fails is replanned and regenerated once with the failure record; a second failure holds the state. The observer records appearance, state, location and world facts, replacing an older fact with the same subject, kind and key. Retrieval supplies one earlier source or endpoint image, excludes images that precede a relevant observed change, and ranks the remaining images by subject and property overlap, then recency.

\subsection{Video Presentation}
\label{impl:sec:transition_video}
Sol-H3 generates $1344\times768$ clips conditioned on the source image, the endpoint that passes the check and the request caption. The frames of the clip present the transition; subsequent operations read the panorama state that the commit installs. Rotations and revisits around the same center render the panorama along a linear yaw and pitch path. \Cref{sec:appendix_prompt_contract} gives the benchmark frame counts, and \cref{sec:appendix_sessions} shows frames from four clips.

The demonstrations also play views at rest as loops. Sol-H3 receives one view of a state as both source and endpoint image, with a caption that names motion in place, such as falling water, drifting mist or turning sails. The loops are computed in advance and lie outside the request pipeline: no check, memory record or benchmark reads them, and the state stays as it is.

\section{Evaluation Protocols}
\label{sec:appendix_eval_contract}
\subsection{Public Recovery Comparison}
\label{sec:public_mbench_definitions}
\label{sec:appendix_recovery}
We rescore the six baselines and \sysname{} on 224 common MBench-A item--condition pairs with one version of the public evaluator, fixed weights and a fixed pose estimator. Invalid scores are excluded without imputation: every metric has 224 valid scores except Matrix-Game 2.0 Epipolar and Reprojection, with 205 and 197, and LingBot-World Reprojection, with 222. This evaluator version differs from the original leaderboard pipeline in its geometric-error temperatures and style formula; the original pose configuration and normalization constants are unavailable for scoring a new method. Rescoring a subset of 30 items per baseline preserves the original six-model ranking in all four environment metrics. The intervals in \cref{tab:main-results} use 20,000 item resamples; blind frame judgments use 10,000 resamples stratified by condition. Object recovery is omitted because its official anchor is absent in 99 of 100 object items; human recovery remains unmeasured.

\begin{figure}[!htb]
\centering
\includegraphics[width=\linewidth]{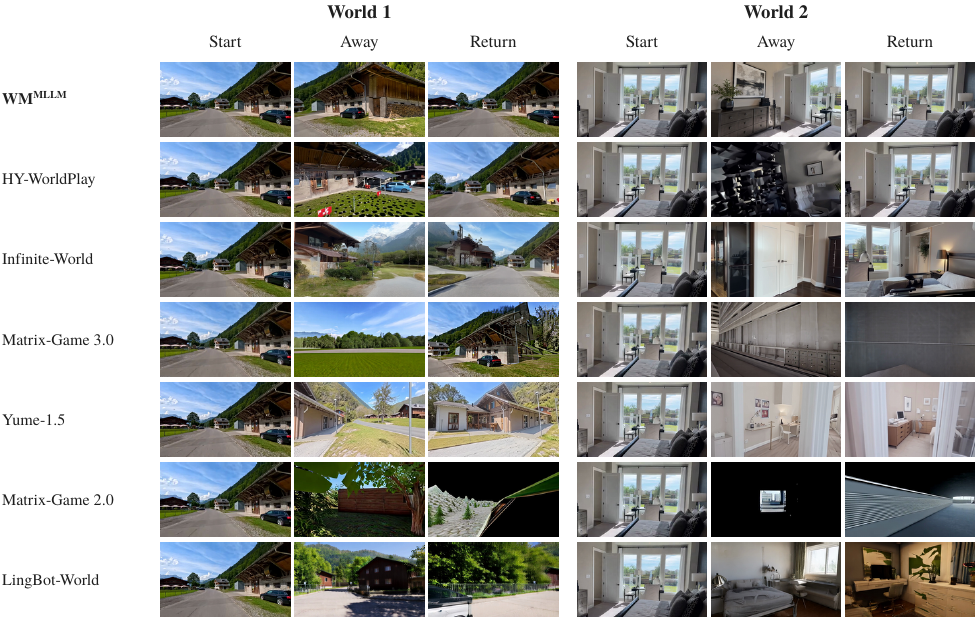}
\caption{\textbf{Camera return and scene recovery.} The baselines can return toward the starting camera while changing the scene. The columns show the start frame, the frame at the farthest turn and the final frame of all seven methods on two MBench-A worlds.}
\label{fig:baseline-gallery-p1}
\end{figure}

\label{sec:appendix_baseline_gallery}
A complete camera round trip turns at least $15^\circ$ in the direction of the request during the clip's first half and ends within $10^\circ$ of the initial orientation. The methods cover different turn sizes: the median outbound yaw is $43^\circ$ for \sysname{} and $82^\circ$ to $173^\circ$ for the baselines. The angular criterion measures camera motion, while the return images in \cref{fig:baseline-gallery-p1} show whether the starting content is recovered.

\label{sec:appendix_env_scorer}
With outbound thresholds of $10^\circ$, $15^\circ$ and $20^\circ$ and return thresholds of $5^\circ$, $10^\circ$ and $15^\circ$, \sysname{} completes the same 191 of 224 items in all nine settings, and the best baseline completes 54 to 158. The full recovery run of \sysname{} has 454 segments, of which 401 render the panorama state geometrically. This reuse contributes to the recovery scores.

\subsection{WBench and Benchmark Run}
\label{sec:appendix_scorers}
We score WBench~\cite{wbenchpaper} with its official metrics and aggregation across quality, setting, interaction, consistency and physical compliance. We replace Doubao-Seed-2.0-lite with Qwen3-VL-8B-Instruct for the fields that a judge model scores and therefore report \sysname{} alone in \cref{tab:wbench-local}. The full inventory combines 158 navigation items with 131 extension items.
\begin{table}[tbp]
\centering
\small
\setlength{\tabcolsep}{3pt}\renewcommand{\arraystretch}{1.15}
\caption{\textbf{WBench scores.} \sysname{}'s highest dimension is consistency in both WBench inventories; its lowest dimension is setting adherence on the navigation items and interaction on the full inventory. Dimension scores are on the 0--100 scale that WBench reports; fields scored by a judge model use Qwen3-VL-8B-Instruct.}
\label{tab:wbench-local}
\begin{tabularx}{\linewidth}{L{.27\linewidth}*{6}{R}}
\tabletoprule
\tablehead
Inventory & Quality & Setting & Interaction & Consistency & Physical & Average \\
\midrule
Navigation (158 items) & 75.6 & 64.7 & 72.5 & 88.5 & 70.7 & 74.4 \\
Full inventory (289 items) & 74.5 & 64.9 & 62.5 & 87.8 & 70.4 & 72.0 \\
\bottomrule
\end{tabularx}
\end{table}

\label{sec:appendix_prompt_contract}
The separate benchmark run used for mechanism diagnostics contains the 158 WBench navigation items and 100 MBench-A causal items, executed in 817 segments. It contains 309 endpoints that pass the verifier, 149 geometric rotations, 46 revisits, 280 failure holds and 33 holds at initialization. The MLLM roles use GPT-6.1-sol and images use GPT-Image 2.5~\cite{gptimage25} with fallback disabled. WBench keeps each clip's 124 native frames. MBench-A resamples a 311-frame sequence to 300 frames and joins two action segments into 600 frames. Both use $1344\times768$ at 24\,fps. Initialization reproduces the input view exactly in all 246 items that obtain a panorama.

\section{Component Studies}
\subsection{Support Composition}
\label{legacy:sec:experiments}
\label{legacy:sec:exp_scope_editing}
\label{sec:appendix_compbench}
The CompBench composition study fixes all 2,709 GPT-Image 2 proposals across three composition rules. Direct editing keeps the whole proposal. Threshold paste copies proposal pixels whose mean RGB change exceeds 25. \sysname{} selects a support and composites onto the source. We score all arms with one CompBench evaluator, average subtasks equally, and bootstrap paired examples and scene clusters. The other system studies use GPT-Image 2.5. The paired LC-I differences of \sysname{} to direct editing, averaged without intermediate rounding, are $+0.0131$ $[0.0119,0.0143]$, $+0.0388$ $[0.0318,0.0463]$ and $+0.0175$ $[0.0122,0.0237]$ on the local, multi-object and implicit-reasoning groups, and to threshold paste $+0.0172$ $[0.0151,0.0192]$, $+0.0130$ $[0.0079,0.0185]$ and $+0.0291$ $[0.0202,0.0380]$ (95\% intervals).

Claude Opus 5.5 judges 150 CompBench proposals, 50 per group, from the source, instruction and three outputs in random order without arm names. The estimates weight groups and subtasks equally with scene standardization. Completion without other changes is a composite outcome computed after judging. \Cref{fig:ablation-judged} shows both instruction completion and this preservation criterion.

\label{sec:appendix_support_timing}
The study of support timing uses the first proposals of all 112 local edits in the benchmark run, including 15 that end in failure holds. It compares the whole proposal, the planner's polygons intersected with its box before generation, and the selector's support after generation. Registration, dilation and feathering are fixed across the two supports. The same blind judge assesses the three outputs in random order. Edits receive equal weight, with 95\% intervals from 20,000 bootstrap draws over item clusters. The difference between the two supports is 48.2 points (95\% interval 37.9 to 58.3).

\subsection{Observation Memory and Lookup by Revision}
\label{sec:appendix_session_counts}
The study of observation memory pairs 20 eight-turn sessions on ten worlds without video. Both arms keep panoramas, the scene graph and lookup by revision; one removes the observer facts and the retrieval of earlier images. Scripts alternate physical edits, departures, returns and dependent actions. An independent judge labels turns blind to arm, counting failures and censored turns against the denominators of the plan. Both arms complete 7 of 20 sessions and preserve the edit on 32 of 40 returns. Removing observation memory produces no substantial decline in this study.

\label{sec:appendix_revision_lookup}
We replay all 198 view lookups of the benchmark run with the physical revision removed from the lookup and rejection rule, retaining the other checks and overwrite order. This permits 61 lookups to return an older revision; 60 of those views differ from the current state at the same camera.

\subsection{Multi-Turn Components}
\label{sec:appendix_b1}
The multi-turn component study fixes 20 new eight-turn scripts on ten worlds before its main run and excludes its pilot. Each script combines a reversible local edit, relighting, departures and returns, and actions that depend on earlier edits. A missing or incorrect departure fails its dependent chain. Each configuration has 160 planned turns, including 60 dependent turns and 40 returns.

The variant with the minimal planner prompt replaces the operation planner's system message with a description of the tools and coordinates. The variant with the video frame as next state installs the last frame of the video into the prior panorama, updating the current view and next observation while retaining offscreen content. All three configurations allow two endpoint proposals. Roles use GPT-6.1-sol at high reasoning effort, images use GPT-Image 2.5~\cite{gptimage25}, and Sol-H3 generates 124 frames at 24\,fps. Rotations and eligible returns use CPU rendering.

Claude Opus 5.5 judges eight frame samples per clip and the next observation, blind to the configuration and the outcome of the endpoint check. The primary outcomes are session completion and the completion of dependent turns; each requires all applicable action, state, return and constraint labels to succeed. Holds, invalid media, missing labels and requests that were never dispatched count against the denominators of the plan. Intervals use 20,000 bootstrap draws over sessions, paired across configurations, with world clustering as a sensitivity analysis. The full system leads each variant by 5.0 points in session completion (95\% interval 0.0 to 15.0). In the completion of dependent turns it leads the variant with the minimal planner prompt by 1.7 points ($-18.3$ to 20.0) and the variant with the video frame as next state by 3.3 points ($-13.3$ to 18.3). All primary differences have 95\% intervals that include zero; these runs do not establish an improvement from either component.
\begin{table}[tbp]
\centering
\small
\setlength{\tabcolsep}{5pt}\renewcommand{\arraystretch}{1.15}
\caption{\textbf{Multi-turn component study.} The three configurations differ by at most 8 of 160 instructions, 2 of 60 dependent turns and 1 of 20 sessions. Entries are successes over all instructions, dependent turns and eight-turn sessions in the scripts, in the primary analysis.}
\label{tab:b1-outcomes}
\begin{tabular}{lccc}
\tabletoprule
\tablehead
Configuration & \shortstack{Instructions\\completed $\uparrow$} & \shortstack{Dependent turns\\completed $\uparrow$} & \shortstack{Sessions\\completed $\uparrow$} \\
\midrule
\oursrow Full system & 67/160 & 18/60 & 1/20 \\
Minimal planner prompt & 59/160 & 17/60 & 0/20 \\
Video frame as next state & 66/160 & 16/60 & 0/20 \\
\bottomrule
\end{tabular}
\end{table}

On five of the scripts, the full system and a variant that allows one proposal and checks only scene identity and endpoint integrity each complete 23 of 40 turns. This comparison changes the retry policy as well as the check. A secondary analysis replaces 15 session runs blocked by a local budget guard with later reruns, while the primary analysis retains the original runs. The comparison with the minimal planner prompt favors the full system on turn completion in that secondary analysis; session completion is unchanged.

\section{Runtime and Limitations}
\paragraph{Runtime.}
\label{sec:appendix_cost_contract}
The time for state processing runs from segment start through planning, endpoint construction, checks and observation. Sol-H3 time covers clip generation; the time to first playback runs from the request to the first playable video. \Cref{sec:exp_cost} reports timings and hardware.

\paragraph{Endpoint and benchmark limits.}
\label{sec:appendix_judge}
A blind judge labels as complete 190 of the 309 endpoints that pass the verifier and 83 of the 276 that the verifier rejects. Because each request starts from the endpoint of the preceding one, the next segment completes in 100 of 153 cases after a correct endpoint commits and in 25 of 95 after an incorrect one.

\label{sec:appendix_locality}
All 97 local edits that pass the verifier leave every panorama pixel outside their support unchanged. The judge finds that the region that changes matches the request in 92 of these edits; 36 include minor changes beyond the request, all inside the support.

The benchmark run pools original completions and reruns after debugging; reasoning effort and scoring hardware vary across subsets.

\paragraph{Seams and geometry.}
\label{sec:appendix_seams_media}
Blind judgments across 36 directions, weighted by the cosine of pitch, give a usable panorama coverage of 87.6\% (95\% interval 85.7\% to 89.4\%). Some directions contain seams or structures that appear twice. A panorama around one center lacks parallax and metric 3D geometry; translation and unseen surfaces require generation.

\section{Capability Comparison Sources}
\label{sec:appendix_comparison_sources}
The capability comparison in \cref{tab:capabilities} follows the versions that we cite and the interfaces that they document. An empty circle means that the source does not describe the capability.

LingBot-World v2 describes object control with a VLM and SAM and recall within its context window, without an interface for appearance editing. WorldDirector describes planning of object and camera trajectories by a language model and identity memory for dynamic objects; object editing and exact scene return lie outside the interface it describes. Alaya-EVOKE describes explorer and chunk prompts, coverage of its geometry bank and revision branches, without instance editing. ActWorld describes object actions driven by text and a memory of events and objects with constant capacity, without appearance editing inside a mask. LongLive describes prompt recaching, a frame sink and a short window; its evaluations measure temporal consistency and prompt transitions.

Yang et al.'s WorldGPT describes text targeting and masks in a workflow from endpoint images to video, without recurrent scene retrieval; it uses ChatGPT for prompt enhancement, and the source does not describe visual input to its planner. Marble describes regional, object and global style editing over persistent 3D assets and revisits of persistent 3D state through camera keyframes or an origin, without exact RGB recovery; a controller and agent memory are unspecified. WorldWeave describes multimodal scene planning, geometric additions that pass checks and persistent structure, without user edits to objects already in the scene or exact RGB return. StateFlow describes a persistent 3D state that it renders again and object edits that a VLM plans on this state; its update procedure does not describe checking generator edits before installation or exact RGB replay. WorldAgents describes view expansion that a VLM directs, checks in two stages before it adds a view and a renderable Gaussian scene, without object editing that a user targets; its return entry applies to a reconstruction that has not changed, rendered with the same camera and rendering profile.

The \sysname{} entry follows the selection and scene storage code of the application and its interfaces for endpoint planning and checking.
\FloatBarrier

\section{Qualitative Examples}
\label{sec:appendix_sessions}
\label{sec:appendix_gallery}
\begin{figure}[tbp]
\centering
\includegraphics[width=\linewidth]{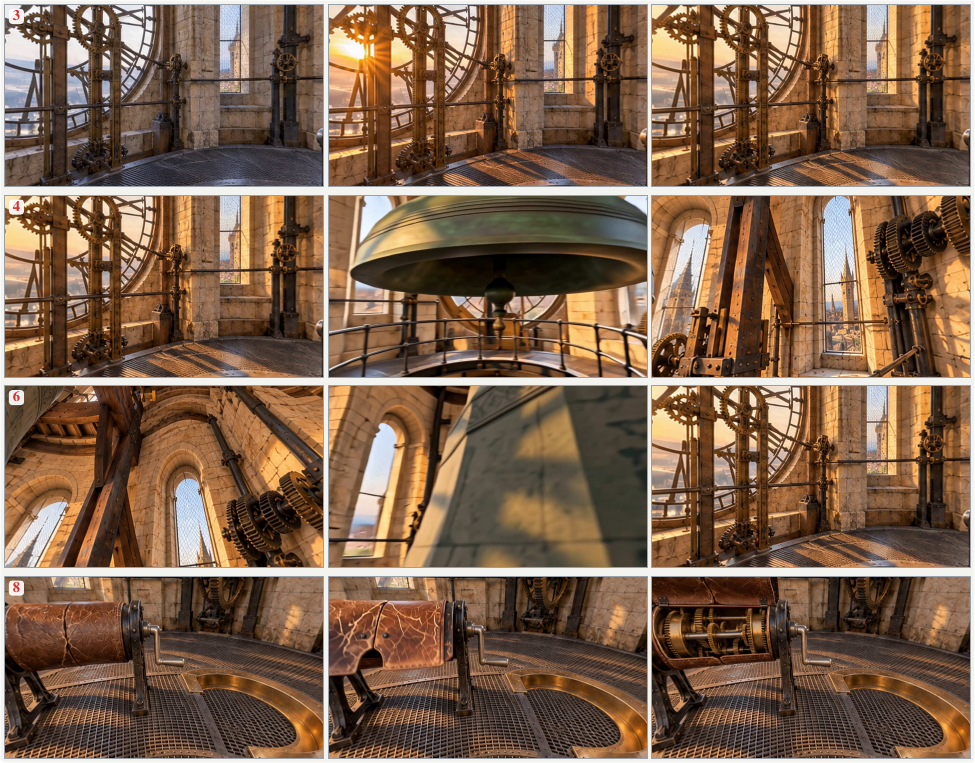}
\caption{\textbf{Transition videos of one session.} Sol-H3 presents each turn as one clip that starts at the view before the turn and ends at the view that the commit installs, and the clip does not write the state. Rows: turns~3, 4, 6 and~8 of the session in \cref{fig:session}: golden evening light through the clock face (3), a walk along the catwalk to the bell (4), a walk back to the start facing the clock face (6) and opening the leather cover of the drum (8); the clip of turn~6 ends at the view of turn~3 in memory. Columns: the first, middle and last of the 124 frames of each clip.}
\label{fig:gallery-overview}
\end{figure}

\Cref{fig:teaser} uses the system of the multi-turn component study (\cref{sec:appendix_b1}) for one world and earlier example sessions for the other three. Those earlier sessions use the region that the planner predicts for local edits and a $110^\circ$ presentation field of view, and they include repairs made before release, so they do not measure success at the first attempt. \Cref{fig:method} uses one MBench-A item and turns 3--6 of a session that runs the same system as that study; this session is not among the scripts of the study. \Cref{fig:session} shows six of its eight turns, and \cref{fig:gallery-overview} shows the Sol-H3 videos of four of them.

\subsection{Global Changes}
\label{sec:appendix_global}
\begin{figure}[tbp]
\centering
\includegraphics[width=\textwidth]{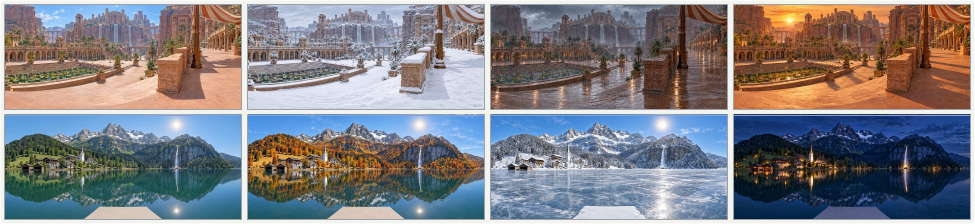}
\caption{\textbf{Weather, season and light.} Each change rewrites the whole panorama and keeps the layout of the scene. Top, from the left: a canyon city at the start, after snow, after rain and at sunset. Bottom: an alpine lake at the start, in autumn, in winter with a frozen lake and at dusk. Front region of each panorama.}
\label{fig:global-changes}
\end{figure}

A global change replaces the panorama and retires the earlier patches. \Cref{fig:global-changes} shows six global changes of weather, season and light, three from each of two start states; the endpoint verifier passes each of them.

\subsection{Long Sessions}
\label{sec:appendix_long_sessions}
\begin{figure}[tbp]
\centering
\includegraphics[width=\textwidth]{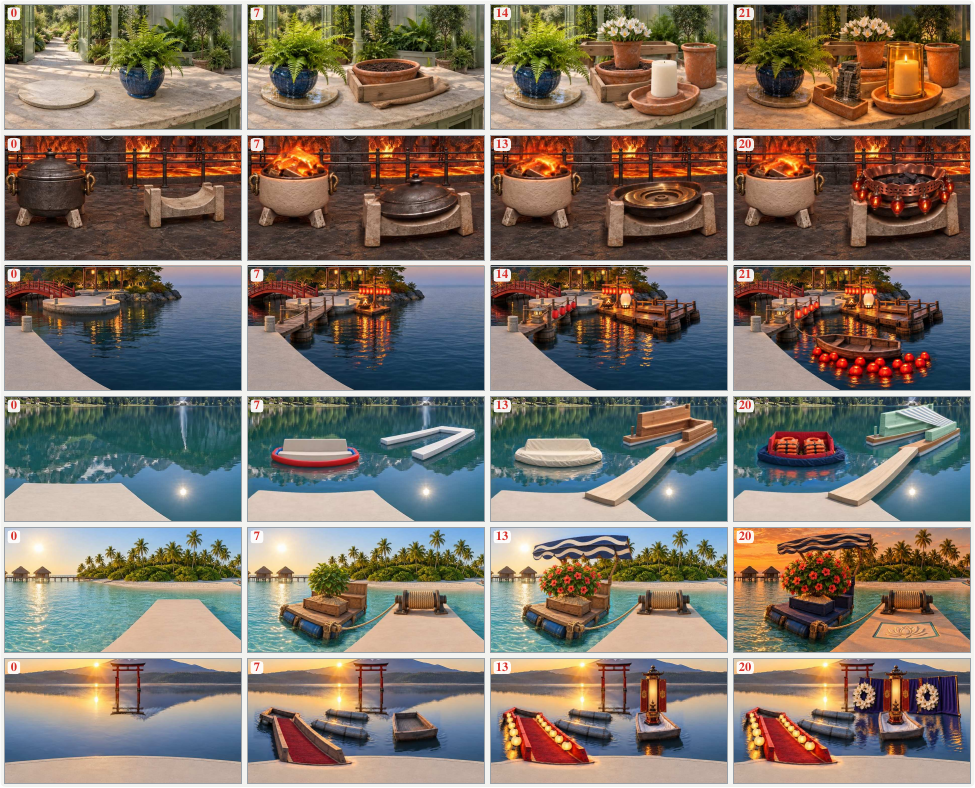}
\caption{\textbf{Six long sessions.} In every world, each request builds on the state that the previous request committed. Each row shows the region that the local requests of one world change, at the start and after about one third, two thirds and all of its requests: a conservatory (21 requests, the session of \cref{fig:play}), a volcanic foundry, a lantern harbor, an alpine lake, a lagoon pier and a torii lake. The last requests of the conservatory and the lagoon pier relight the whole panorama.}
\label{fig:sessions-six}
\end{figure}

\Cref{fig:sessions-six} shows six worlds with 20 or 21 sequential requests each, and \cref{tab:session-requests} lists the requests of the conservatory session of \cref{fig:play}. Each request starts from the state that the previous request committed. In the conservatory session, requests 1--20 change pixels only inside one box that covers 10.7\% of the panorama, so the panorama after request 20 equals the start panorama outside the box; request 21 relights the whole panorama. This session stores $2048\times1024$ panoramas and writes each local edit as a $1344\times768$ patch with a horizontal field of view between $90^\circ$ and $120^\circ$; GPT-6.1-sol runs at reasoning effort xhigh, and images use GPT-Image 2.5. An agent writes each request from the current state, and candidates that fail the checks are generated again, so these sessions do not measure success at the first attempt.

\begin{table}[htbp]
\centering\footnotesize
\setlength{\tabcolsep}{4pt}\renewcommand{\arraystretch}{1.05}
\captionsetup{singlelinecheck=false}\caption{\textbf{Requests of the conservatory session.} \Cref{fig:play} shows the state after each request.}
\label{tab:session-requests}
\begin{tabularx}{\linewidth}{rY}
\tabletoprule
\tablehead
\# & Request \\
\midrule
1 & Slide the blue fern planter along the bench onto the stone pad, then lay a flat hessian sack where it stood. \\
2 & Place one broad terracotta irrigation bowl on the potting bench beside the stone-pad fern. \\
3 & Fill the terracotta irrigation bowl with water on the bench beside the stone-pad fern. \\
4 & Pour the terracotta bowl water into the blue fern planter until it overflows onto the stone pad. \\
5 & Move the drained terracotta bowl onto the front-left corner of the fixed bench hessian. \\
6 & Build one compact wooden nursery trough around the empty terracotta bowl at the hessian's front-left corner. \\
7 & Fill the terracotta bowl with dark soil inside the nursery trough on the bench hessian. \\
8 & Place one tall cylindrical terracotta pot of dormant white bulbs into the soil-filled bowl inside the wooden trough. \\
9 & Bloom the dormant white bulbs into a broad cluster of short white crocuses in their terracotta pot inside the nursery trough. \\
10 & Build one compact low timber nursery stand with a deep flat reservoir shelf on the bench behind the flowering pot. \\
11 & Place one tall wide water-filled terracotta reservoir on the nursery stand shelf beside the white flowers. \\
12 & Move the filled terracotta reservoir from the nursery stand shelf onto the stone bench beside the flowering trough. \\
13 & Build one shallow clay basin with a broad dry candle pedestal on the stone bench in front of the nursery trough beside the filled reservoir. \\
14 & Place one large white pillar candle on the clay basin broad dry pedestal. \\
15 & Put one broad clear glass chimney around the white pillar candle, resting on the clay basin dry pedestal. \\
16 & Light the pillar candle inside the glass chimney, casting warm candlelight onto the clay basin. \\
17 & Build one low brass lantern stand around the lit chimney on the clay basin broad dry pedestal. \\
18 & Place one rigid amber shade on the lantern stand around the candle chimney above the clay basin. \\
19 & Place one narrow empty rectangular ceramic water garden with a raised dry pocket between the fern pad and lantern basin, leaving stone gaps on both sides. \\
20 & Place one compact running stone fountain on the main basin floor of the ceramic water garden, clear of its raised dry planting pocket. \\
21 & Evening falls outside the glass and the amber lantern lights the potting bench. \\
\bottomrule
\end{tabularx}
\end{table}

\end{document}